%% file: top.tex
\documentclass[10pt,twocolumn,letterpaper]{article}
\usepackage{multicol}

\usepackage{iccv}
\usepackage{times}
\usepackage{epsfig}
\usepackage{graphicx}
\usepackage{amsmath}
\usepackage{amssymb}

\usepackage{times} 

\usepackage{float}
\usepackage{booktabs}
\usepackage{caption}
\usepackage{multirow}
\usepackage{xcolor}

\mathchardef\mhyphen="2D
\usepackage[utf8]{inputenc}
\usepackage{comment}
\usepackage{tabularx,colortbl}
\usepackage{xparse,mathtools}
\usepackage{diagbox}
\usepackage{adjustbox,lipsum}

\usepackage{url}
\usepackage[numbers,sort,compress]{natbib}

\usepackage[accsupp]{axessibility}  

\usepackage{tabulary,overpic}

\newlength\savewidth

\usepackage{siunitx}
\input{shortcuts}

\usepackage{amssymb}
\usepackage{pifont}
\newcommand{\cmark}{\ding{51}}%
\newcommand{\xmark}{\ding{55}}%

\newcommand\zhongkai[1]{\textcolor[rgb]{1,0,1}{#1}}
\newcommand\lei[1]{\textcolor[rgb]{0,0,1}{#1}}

\colorlet{lr}{red!70!white}
\colorlet{lb}{cyan!70!white}
\colorlet{lo}{orange!70!white}
\colorlet{lm}{magenta!70!white}

\definecolor{lb}{HTML}{FFEFCC}
\definecolor{lo}{HTML}{CCFFEF}
\definecolor{lm}{HTML}{CDCCFF}
\usepackage[pagebackref=true,breaklinks=true,letterpaper=true,colorlinks,bookmarks=false]{hyperref}

\iccvfinalcopy 

\def\iccvPaperID{2896} 

\ificcvfinal\fi

\begin{document}






\title{XVO: Generalized Visual Odometry via Cross-Modal Self-Training}

\author{Lei Lai$^*$ \quad Zhongkai Shangguan$^*$ \quad Jimuyang Zhang \quad Eshed Ohn-Bar\\
Boston University\\
{\tt\small \{leilai, sgzk, zhangjim, eohnbar\}@bu.edu}
}

\maketitle

\newcommand\blfootnote[1]{%
\begingroup
\renewcommand\thefootnote{}\footnote{#1}%
\addtocounter{footnote}{-1}%
\endgroup
}

\blfootnote{$^*$ Equally contributed.}

\ificcvfinal\thispagestyle{empty}\fi

\input{sec_abstract}
\input{sec_intro}
\input{sec_related}

\input{sec_method}

\input{sec_analysis}

\input{sec_conclusion}

{
\small
\bibliographystyle{ieee_fullname}
\bibliography{egbib}
}

\clearpage
\onecolumn
\begin{center}
      {\Large \bf Supplementary Material for \\ XVO: Generalized Visual Odometry via Cross-Modal Self-Training \par}
     \vspace*{16pt}
      \vskip .5em
      \vspace*{12pt}
\end{center}

\setcounter{section}{0}

\input{sec_supp}

\end{document}

%% file: shortcuts.tex
\newcommand{\bA}{\mathbf{A}}

\newcommand{\bD}{\mathbf{D}}

\newcommand{\bF}{\mathbf{F}} 

\newcommand{\bI}{\mathbf{I}}

\newcommand{\bR}{\mathbf{R}}
\newcommand{\bS}{\mathbf{S}}
\newcommand{\bt}{\mathbf{t}}

\newcommand{\bx}{\mathbf{x}}
\newcommand{\by}{\mathbf{y}}

\newcommand{\btheta}{\boldsymbol{\theta}}

\newcommand{\bPsi}{\boldsymbol{\Psi}}

\newcommand{\cD}{\mathcal{D}}

\newcommand{\cL}{\mathcal{L}}

\newcommand{\cR}{\mathcal{R}}

\makeatletter
\DeclareRobustCommand\onedot{\futurelet\@let@token\@onedot}
\def\@onedot{\ifx\@let@token.\else.\null\fi\xspace}
\def\eg{e.g\onedot} 
\def\ie{i.e\onedot}

\def\etal{et~al\onedot}

\makeatother

\newcommand{\boldparagraph}[1]{\vspace{0.2cm}\noindent{\bf #1:}}

\definecolor{darkgreen}{rgb}{0,0.7,0}

\newcommand{\red}[1]{\noindent{\color{red}{#1}}}

%% file: sec_abstract.tex

\begin{abstract}

We propose XVO, a semi-supervised learning method for training generalized monocular Visual Odometry (VO) models with robust off-the-self operation across diverse datasets and settings. In contrast to standard monocular VO approaches which often study a known calibration within a single dataset, XVO efficiently learns to recover relative pose with real-world scale from visual scene semantics, \ie, without relying on any known camera parameters. We optimize the motion estimation model via self-training from large amounts of unconstrained and heterogeneous dash camera videos available on YouTube. Our key contribution is twofold. First, we empirically demonstrate the benefits of semi-supervised training for learning a general-purpose direct VO regression network. Second, we demonstrate multi-modal supervision, including segmentation, flow, depth, and audio auxiliary prediction tasks, to facilitate generalized representations for the VO task. Specifically, we find audio prediction task to significantly enhance the semi-supervised learning process while alleviating noisy pseudo-labels, particularly in highly dynamic and out-of-domain video data. Our proposed teacher network achieves state-of-the-art performance on the commonly used KITTI benchmark despite no multi-frame optimization or knowledge of camera parameters. Combined with the proposed semi-supervised step, XVO demonstrates off-the-shelf knowledge transfer across diverse conditions on KITTI, nuScenes, and Argoverse without fine-tuning. 

\end{abstract}

%% file: sec_intro.tex
\section{Introduction}
\label{intro}
\begin{figure}[!t]
    \centering
    \includegraphics[trim={0cm 1.5cm 6.5cm 0cm},clip,width=3.7in]{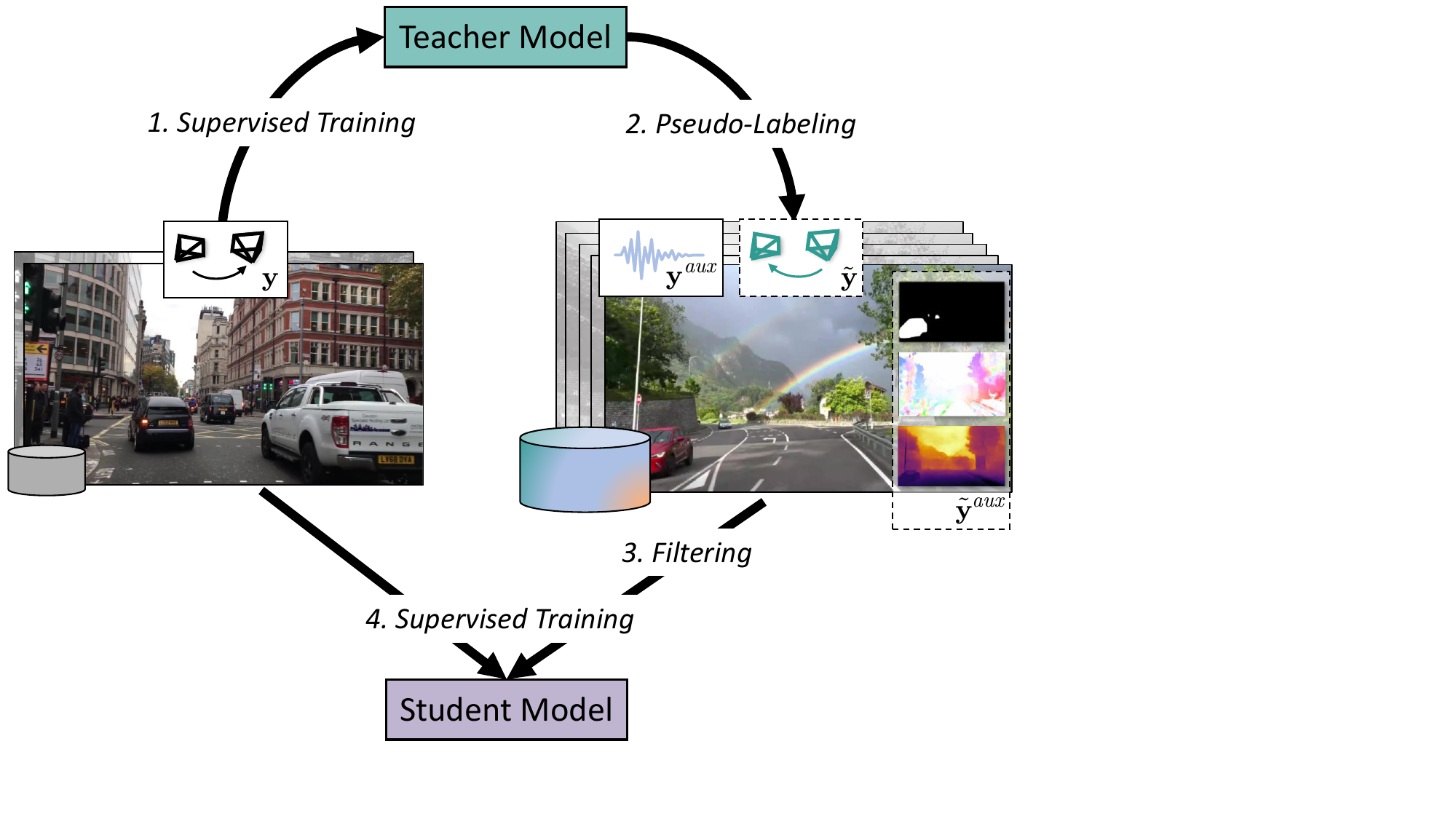}
    \vspace{-0.5cm}
    \caption{\textbf{Learning General-Purpose Monocular Visual Odometry (VO) Models from Multi-Modal and Pseudo-Labeled Videos.}
   Our proposed XVO framework first trains an ego-motion prediction \textit{teacher model} over a small initial dataset, \eg, nuScenes~\cite{caesar2020nuscenes}. We then expand the original dataset through pseudo-labeling of in-the-wild videos. Motivated by how humans learn general representations through observation of large amounts of multi-modal data, we employ multiple auxiliary prediction tasks, including segmentation, flow, depth, and audio, as part of the semi-supervised training process. Finally, we leverage \textit{uncertainty-based filtering} of potentially noisy pseudo-labels and train a robust student model. 
}
\vspace{-.45cm}
    \label{fig:flowchart}
\end{figure}

Monocular Visual Odometry (VO) methods for recovering ego-motion from a sequence of images have mostly been studied within a \textit{restricted scope}, where a single dataset, such as KITTI~\cite{Geiger2012CVPR}, may be used for both training and evaluation under a fixed pre-calibrated camera~\cite{yin2018geonet,yang2020d3vo,li2018undeepvo,mahjourian2018unsupervised,zhan2020visual,zhao2020towards,zhou2017unsupervised,iyer2018geometric,rockwell20228,zou2018df}. However, very few studies have analyzed the task of \textit{generalized VO}, \ie, relative pose estimation with real-world scale across differing scenes and capture setups.

 For instance, consider an autonomous robot or vehicle deployed at a large scale. The robot is highly likely to encounter environments for which no ground truth ego-motion data was previously collected. In such novel settings, current VO methods will quickly exhibit poor ego-motion estimation and drift~\cite{wang2021tartanvo,li2021generalizing,gordon2019depth,li2019sequential,zou2018df,mur2017orb,engel2014lsd,engel2017direct}. Moreover, our robot may be required to adjust its camera setup over its lifetime (\eg, to a new camera) or leverage data from a fleet of robots with varying or perhaps unknown camera configurations. Yet, existing VO methods generally assume carefully calibrated camera parameters during training~\cite{wang2021tartanvo,rockwell20228,yang2020d3vo,zou2018df,mur2017orb,engel2014lsd,engel2017direct}. 
Specifically, to simplify the ill-posed monocular pose recovery task, researchers often resort to relying on knowledge of the camera intrinsic parameters to incorporate various geometric or photometric consistency-based mechanisms~\cite{li2018undeepvo,mahjourian2018unsupervised,zhan2020visual,zhao2020towards,zhou2017unsupervised,iyer2018geometric,yin2018geonet,teed2022deep}. In this work, we do not make such an assumption as we are concerned training VO models that can learn from and operate under diverse unconstrained videos in the wild. Specifically, we pursue an orthogonal direction to prior work based on semi-supervised learning and explore more scalable and camera-agnostic deployment settings. 

Our key hypothesis is that neural network models can learn to circumvent issues related to pose and scale ambiguity in generalized VO settings through observation of ample amounts of diverse scene and motion video data. This approach is motivated by humans' ability to flexibly estimate motion in arbitrary conditions through a general understanding of salient scene properties (\eg, object sizes)~\cite{pitzalis2013selectivity}. This general understanding is developed over large amounts of perceptual data, often multi-modal in nature~\cite{smith2005development,smith2005development,piaget1952origins}. For instance, cross-modal information processing between audio and video has been shown to play a role in spatial reasoning and proprioception~\cite{maslen2022hearing,o2008seeing,marian2021cross,watkins2006sound,radeau1994auditory}. Indeed, collected online videos often have audio, which can be used as a further source of cross-modal supervision. As further discussed in Sec.~\ref{sec:method}, we find ambient audio to correlate at times with scenarios where monocular VO tends to fail, such as determining ego-speed when the vehicle is stopped at a dense intersection or as context for the current traffic scenario when estimating translation (\eg, highway driving). Extracting information related to flow, segmentation, or depth can also further guide learning generalized representations. 
To fully explore the utility of self-training VO models, we analyze a unified multi-modal framework and its impact on guiding semi-supervised VO learning from large amounts of unconstrained sources. 

As far as we are aware, we are the first to study the feasibility of self-training for direct, calibration-free, ego-motion pose regression with an absolute real-world scale. Specifically, we find that incorporating additional modalities via simple multi-task learning can significantly enhance model robustness and generalization. When paired with an uncertainty-based filtering module, we achieve state-of-the-art VO performance with a single broadly usable model which we validate for the autonomous driving use-case. Moreover, our training and inference is highly efficient as the auxiliary learning formulation does not alter the two-frame input, \ie, in contrast to methods relying on extracting rich intermediate representations~\cite{wang2021tartanvo,zhao2020towards,behl2020label,yin2018geonet}. We demonstrate state-of-the-art results on KITTI using the proposed two-frame VO model structure without requiring elaborate long-term memory, computationally expensive iterative refinement steps, or knowledge of camera parameters. Our code is available at~\url{https://genxvo.github.io/}. 

%% file: sec_related.tex
\section{Related Work}
\label{sec:related}
\vspace{-0.2cm}
\boldparagraph{Monocular Visual Odometry}
Despite recent advances, both geometrical and learning-based VO approaches are still mostly evaluated over limited datasets under similar training and testing conditions~\cite{mur2015orb,campos2021orb,vakhitov2018stereo,wang2020feature,esfahani2019aboldeepio,piao2019real,kendall2015posenet,wang2017deepvo}. 
For instance, training and evaluation are both conducted on KITTI~\cite{Geiger2012CVPR}, which contains a fixed camera setup with limited diversity and density of scenarios~\cite{wang2021tartanvo}.
More recently, learning-based approaches leveraging unsupervised learning for VO~\cite{li2018undeepvo, yin2018geonet, zhou2017unsupervised, ranjan2019competitive, li2019sequential, zhan2018unsupervised}, have shown state-of-the-art performance on KITTI. Notably, UnDeepVO~\cite{li2018undeepvo} utilizes stereo imagery for training to recover real-world scale without the need for labeled datasets. GeoNet~\cite{yin2018geonet} combines depth, optical flow, and camera pose to holistically learn a VO prediction model. TartanVO~\cite{wang2021tartanvo} conditions the VO model on the intrinsic parameters in order to achieve robust generalized performance. However, the aforementioned methods all require known camera intrinsics~\cite{li2018undeepvo, yin2018geonet} in inference resulting in a restricted use-case and cannot leverage data with unknown camera parameters. 
In light of these challenges, our work develops mechanisms to enable VO models to learn from and operate over diverse datasets without known calibration. Specifically, our method leverages semi-supervised and multi-modal learning techniques to learn robust generalized representations for estimating motion and real-world scale. Therefore, our approach is orthogonal to most related methods which emphasize self- and un-supervised learning of models based on warping and consistency tasks which rely on precise camera calibration~\cite{li2018undeepvo,iyer2018geometric,yin2018geonet}.

\boldparagraph{Semi-Supervision for Computer Vision Tasks}
Semi-supervised learning approaches have been extensively studied within the computer vision and machine learning communities~\cite{li2019label, tarvainen2017mean, denis2020consistency, yin2022fishermatch, sohn2020fixmatch, mann2007simple}. However, prior works have not yet focused on the monocular VO task, instead emphasizing object detection~\cite{caine2021pseudo}, semantic segmentation~\cite{wang2022semi}, 3D reconstruction~\cite{xing2022semi}, action recognition~\cite{xing2023svformer} or low-level computer vision tasks~\cite{jeong2019consistency, souly2017semi, miyato2016adversarial,yan2019semi, berthelot2019mixmatch,yang2021survey,carmon2019unlabeled, hendrycks2019using, deng2021improving, peikari2018cluster, mey2022improved}. Consequently, fundamental research questions related to the impacts of improving VO model generalization remain unanswered, \eg, whether semi-supervised learning can be used to enhance reasoning over real-world scale~\cite{lee2013pseudo,wang2021uncertainty,rizve2021defense,zhang2022selfd} and long-tail scenarios~\cite{xworld} or how uncertainty mechanisms can contribute to more robust training from heterogeneous video data. Specifically, we explore the role of multi-task and multi-modal learning in order to improve semi-supervised VO model training.

\boldparagraph{Auxiliary Learning} Our proposed method primarily aims to enhance the performance of a VO model through auxiliary tasks within a semi-supervised learning framework. Auxiliary learning~\cite{liebel2018auxiliary,zhang2014facial, liu2019self} aims to use auxiliary tasks to enhance the performance of the primary tasks. 
It has been effectively employed in diverse domains, including computer vision~\cite{misra2016cross, kokkinos2017ubernet, xu2021leveraging, dadashzadeh2022auxiliary}, natural language processing~\cite{deng2022improving, kung2021efficient, dery2022aang}, and robotics~\cite{ohn2020learning,Nigam-2018-105996, wang2022goal, zhang2021learning, islam2019real, song2021multimodal,zhang2023coaching}. 
Xu~\etal~\cite{xu2021leveraging} applied auxiliary image classification and saliency detection to improve the performance of
the semantic segmentation. Song~\etal~\cite{song2021multimodal} leverages an auxiliary task of velocity estimation to enhance the ability to avoid obstacles of an indoor mobile robot. While several related studies employ auxiliary supervision derived from the ground-truth depth and optical flow~\cite{teed2021droid,teed2022deep}, in this work our goal is to explore the use of such supervision from potentially noisy pseudo-labels, \eg, as regularization for learning robust internal representations for VO.








\boldparagraph{Cross-Modal Learning}
Cross-modal learning is inspired by how biological systems learn by incorporating complementary information from multiple modalities, such as vision, sound, and touch. Prior research in computational cross-modal learning has focused on learning a shared representation space where samples from distinct modalities \ie, image, audio, text, can be aligned~\cite{zhen2019deep, baltruvsaitis2018multimodal,huang2022assister}. Moreover, the addition multiple tasks and modalities have been shown to benefit generalization for various machine perception and learning tasks~\cite{bachmann2022multimae,zamir2018taskonomy,zamir2020robust,zamir2018taskonomy,sax2018mid,huang2022assister}. For instance, audio generation~\cite{di2021video, tan2020automated, zhou2018visual}, image captioning~\cite{xu2015show,li2020oscar}, speech recognition~\cite{afouras2018deep,serdyuk2021audio}, navigation~\cite{chen2020soundspaces}, and multimedia retrieval~\cite{gasser2019multimodal, chang2015text, ionescu2022overview} have all shown improved performance due to cross-modal training. However, such studies tend to focus on simplified domains, \eg, restricted acoustic or haptic environments, whereas we analyze dense and dynamic scenes in the wild.

%% file: sec_method.tex
\section{Method}
\label{sec:method}

Our proposed framework comprises three main steps: (1) uncertainty-aware training of an initial (\ie, teacher) VO model (Sec.~\ref{subsec:teacher}); (2) pseudo-labeling with the removal of low-confidence and potentially noisy samples (Sec.~\ref{subsec:filter}); (3) self-training with pseudo-labeled and auxiliary prediction tasks of a robust VO student model (Sec.~\ref{subsec:aux}).

\subsection{Problem Setting}
\boldparagraph{Direct Pose Regression} Our goal is to learn a general function for mapping two observed image frames $\bx{_i}~=~\{\bI_{i-1},\bI_i\}$, with $\bI \in \cR^{W \times H \times 3}$, to a relative camera pose with real-world scale $\by_i=[\bR_i|\bt_i] \in \text{SE(3)}$ with rotation $\bR_i\in \text{SO}(3)$ and translation $\bt_i\in \mathbb{R}^3$. Given a dataset comprising annotated labels of pose ground-truth, $\cD_L = \{(\bx{_i}, \by{_i})\}_{i=1}^N$, learning-based approaches for VO often optimize for a regression loss~\cite{teed2022deep,wang2017deepvo,yin2018geonet,ranjan2019competitive,behl2019pointflownet}. In practice, the direct pose regression task often exhibits drift due to issues with absolute scale ambiguity and compound errors, particularly in cases of dense and dynamic scenes. For instance, small errors in rotation estimation can result in large errors over multiple time steps which impact the evaluation. While we formulate a two-frame regression task, prior methods have relied on longer-term memory in order to improve model robustness~\cite{wang2017deepvo,iyer2018geometric,li2019sequential,xue2019beyond}, however, this comes at a computational and memory cost. Moreover, most monocular methods only produce up-to-scale predictions~\cite{wang2021tartanvo,yin2018geonet,li2021generalizing}, as will be further discussed in Sec~\ref{sec:analysis}. Instead, we rely on a semi-supervised training process to mitigate issues in absolute scale recovery while enabling a simple two-frame model to achieve state-of-the-art results. 

\boldparagraph{Self-Training with Auxiliary Tasks}
In addition to a labeled odometry dataset $\cD_L$, our framework assumes access to a large dataset that is not annotated with respect to the ego-motion task but potentially other complementary tasks that are auxiliary to the main VO task, \ie, $\cD_U = \{(\bx{_i}, \by_{i}^{aux} )\}_{i=1}^M$. Moreover, we assume access to a set of models for generating a \textit{pseudo-labeled} dataset~\cite{rizve2021defense,caine2021pseudo,yan2019semi,lee2013pseudo,zhang2022selfd}, \ie, $\cD_{PL} = \{(\bx{_i},\tilde{\by}_{i}, \by_{i}^{aux}, \tilde{\by}_{i}^{aux})\}_{i=1}^M$ which can be joined with the original dataset $\tilde{\cD}~=~\cD_L~\cup~ \cD_{PL}$ for supervised training (Sec.~\ref{subsec:aux}). We note that this is a practical assumption as there are abundant computer vision models for obtaining various pseudo-labels. As will be discussed in Sec.~\ref{subsec:filter}, these pseudo-labels may be filtered by removing high-uncertainty samples.  
Overall, the cross-modal self-training objective can be defined as
\begin{equation}
    \cL_{xvo} = \cL_{vo}(\by) + \lambda_u \cL_{unc}(\by) + \cL_{aux}(\by^{aux},\tilde{\by}^{aux})
\end{equation}
where $\cL_{vo}$ is a main VO task loss, $\cL_{unc}$ is an uncertainty estimation loss, $\cL_{aux}$ is defined over the auxiliary prediction tasks, and $\lambda_u$ is a scalar hyper-parameter. 
We demonstrate our semi-supervised formulation to benefit various known issues with VO, \eg, improving scale recovery. Moreover, our formulation is kept efficient during inference as it does not alter the two-frame input $\bx$, \ie, in contrast to methods relying on extracting intermediate representations as input, such as flow~\cite{wang2021tartanvo} or depth~\cite{zhao2020towards}. 
Next, we define our network structure and training.

\begin{figure}
    \centering
    \includegraphics[width=3.3in]{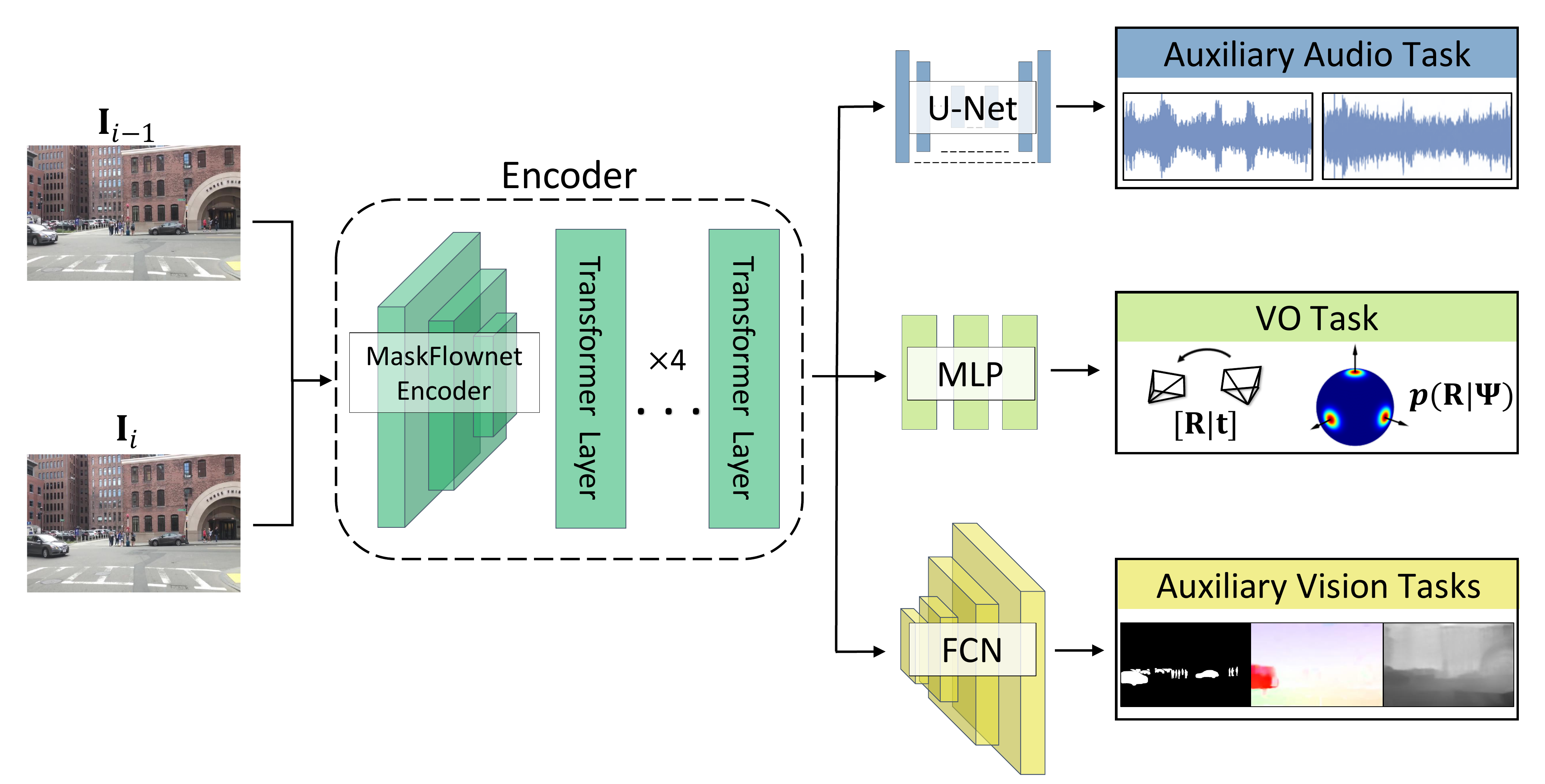}
    \caption{\textbf{Network Architecture.} Our initial teacher model (used for pseudo-labeling and filtering) encodes two concatenated image frames and predicts relative camera pose and its uncertainty. The complete cross-modal architecture leverages a similar architecture but with added auxiliary prediction branches with complementary tasks that can further guide self-training, \eg, prediction branches for audio reconstruction, dynamic object segmentation, optical flow, and depth. 
}
    \label{fig:modelstructure}
\end{figure}

\subsection{Ego-Motion Network Training}
\label{subsec:teacher}

Our approach first trains a direct ego-motion teacher model (shown as the main encoder and middle branch in Fig.~\ref{fig:flowchart}) over the labeled dataset $\cD_L$. To enable learning from an unconstrained video, we do not incorporate any dependency on intrinsic parameters, \ie, either as input~\cite{wang2021tartanvo,engel2014lsd,engel2017direct} or for computing a supervisory loss~\cite{li2018undeepvo,yin2018geonet,teed2022deep,rockwell20228,yang2020d3vo}. We find our network design to provide an efficient but surprisingly strong baseline, matching state-of-the-art on the KITTI benchmark despite no elaborate multi-frame optimization.

\boldparagraph{Encoder} We employ a high-capacity feature extractor for effectively leveraging the rich multi-task supervision in later stages (Sec.~\ref{subsec:aux}). The feature extractor is a MaskFlownet encoder~\cite{zhao2020maskflownet}, which was found to outperform the commonly used PWC-Net~\cite{sun2018pwc,wang2021tartanvo}, followed by four transformer self-attention layers~\cite{dosovitskiy2021an,chen2021transunet}. The patch size is $12\times 16$, with each layer comprising four heads and $256$ hidden parameters. The encoder structure for the initial teacher model and cross-modal student is kept the same.

\boldparagraph{VO Decoder} The VO decoder branch consists of three Fully Connected (FC) layers that regress relative pose $\by~=~[\bR|\bt]$ and an uncertainty estimate for the prediction. The VO task optimizes a Mean Squared Error (MSE) loss over predicted translation $\hat{\bt} \in \cR^3$ and Euler angle rotations $\hat{\btheta} \in \cR^3$,
\begin{equation}
    \cL_{vo} = \|\bt-\hat{\bt}\|_2^2  + \lambda_\theta \|\btheta-\hat{\btheta} \|_2^2
\end{equation}


\boldparagraph{Uncertainty Estimation} To account for the difficulty in the absolute scale pose regression task, we propose to also model prediction uncertainty. We adopt a matrix Fisher distribution~\cite{mohlin2020probabilistic}, which provides a framework for modeling rotation distribution on $\text{SO}(3)$. The probability density function of the matrix Fisher distribution is as follows:
\begin{equation}
p(\bR|\bPsi)=\frac{1}{c(\bPsi)}\exp(tr(\bPsi^\top \bR))
\end{equation}
where $\bPsi\in \mathbb{R}^{3\times3}$ are the distribution parameters, $\bR \in \text{SO}(3)$ is the pose rotation matrix, and $c(\bPsi)$ is a normalization constant~\cite{mardia2000directional}. Given the estimated parameters $\hat{\bPsi}$ we use the negative log likelihood of $\bR$ in the predicted distribution as a loss, \ie, 
\begin{equation}
    \cL_{unc} = -\log(p(\bR|\hat{\bPsi}))
\end{equation}

As a proxy for prediction (\ie, pseudo-label) quality, we find it is sufficient to model uncertainty in rotation prediction, however more elaborate estimation methods can also be used~\cite{rizve2021defense,yang2021st3d,kendall2018multi}. The confidence predictions will be used to remove potentially noisy pseudo-labels prior to the self-training process, as discussed next. 

\subsection{Pseudo-Label Selection}
\label{subsec:filter}
The VO model from Sec.~\ref{subsec:teacher} can be used to obtain pseudo-labels over an unlabeled (\ie, with respect to the main VO task) data $\cD_U$. However, incorrect predictions can introduce noise and heavily degrade model training~\cite{sohn2020fixmatch,rizve2021defense}. Hence, it is crucial to remove low-confidence samples prior to the cross-modal self-training. 

In our regression problem, we measure the confidence of a pseudo-label based on the entropy of the predicted matrix Fisher distribution (\ie, a lower entropy represents increased confidence),
\begin{equation}\label{eq:filter}
      H(p(\bR|\hat{\bPsi}))  <  \tau_{u}
\end{equation}
where we set a fixed threshold $\tau_{u}$ to ensure the network prediction is sufficiently certain to be selected. To generate pseudo-labels, the VO model is tested on out-of-domain data with highly diverse and dynamic scenes. Based on our analysis in Sec.~\ref{sec:analysis}, we find the uncertainty-aware selection mechanism to be crucial for robust self-training irrespective of the auxiliary training tasks.   


\subsection{Self-Training with Auxiliary Tasks}
\label{subsec:aux}

\begin{figure}[!t]
    \centering
    \includegraphics[width=3.25in]{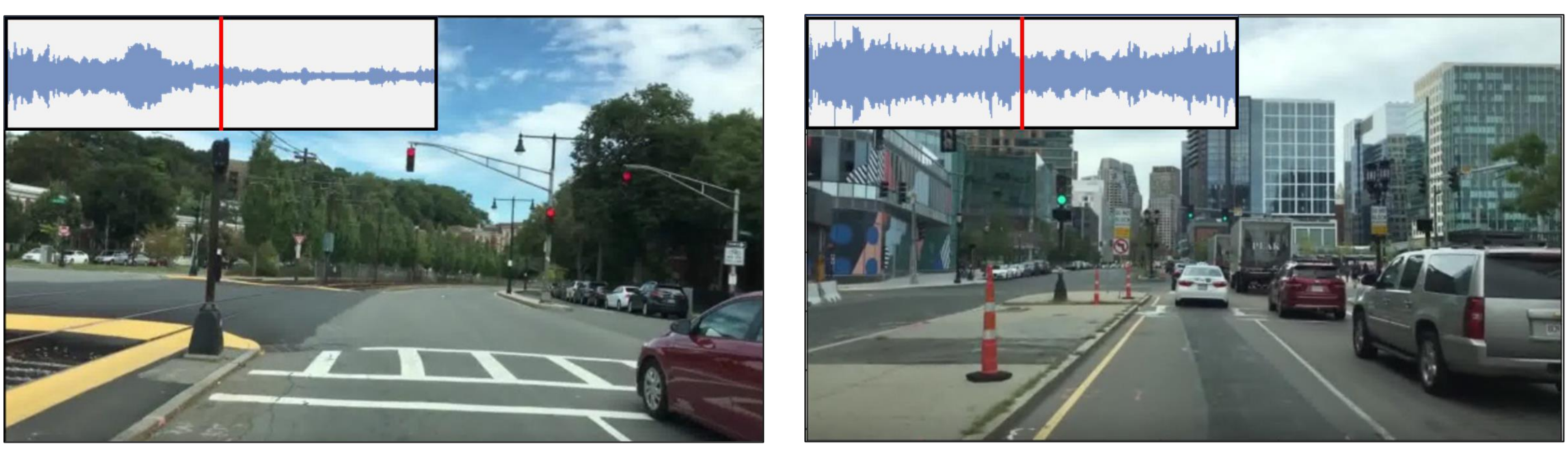}
    \caption{
    \textbf{Illustration of the Importance of Audio.} The frame is consistent with the red arrow marked on the waveform. Left: audio amplitude decreases and maintains a low level when the vehicle is going to wait for traffic lights. Right: audio experiences many ups and downs representing acceleration and brake in a narrow urban area. }
    \vspace{-0.2cm}
    \label{fig:audio}
\end{figure}

To learn effective representations for generalized VO at scale, we propose to incorporate supervision from auxiliary but potentially complementary prediction tasks in addition to the generated VO pseudo-labels on $\cD_U$. The introduced auxiliary tasks regularize the self-training process, particularly in cases where VO pseudo-labels may be inaccurate but other modalities may contain relevant information for reducing ambiguity. Our approach is motivated by the success of multi-task frameworks for computer vision tasks~\cite{bachmann2022multimae,zamir2020robust,zamir2018taskonomy,kokkinos2017ubernet,gordon2019splitnet,zhang2021survey}. However, we emphasize that related studies often leverage high-quality annotated labels and not noisy pseudo-labels based on model predictions. We sought to incorporate useful auxiliary tasks, as unrelated or noisy supervision can impede the learning process and result in a detrimental effect on the main task model. We set the auxiliary labeled task as an audio prediction task $\by^{aux}_i \coloneqq \bA_i \in \cR^{2\times L}$, and the auxiliary pseudo-labeled tasks $\tilde{\by}^{aux}_i \coloneqq [\tilde{\bS}_i, \tilde{\bD}_i, \tilde{\bF}_i ] \in \cR^{W\times H \times C}$ as segmentation, depth, and flow prediction, respectively. Subsequently, we leverage multi-task learning (as shown in Fig.~\ref{fig:audio}) and minimize a loss composed of four terms, 
\begin{equation}
 \cL_{aux} = \lambda_a \cL_{audio} + \lambda_s \cL_{seg} + \lambda_f \cL_{flow}  + \lambda_d \cL_{depth}
\end{equation}
over the entire dataset $\tilde{\cD}$. We note that we drop the explicit label source to avoid clutter. Next, we define each term and corresponding decoder. We empirically observe the additional tasks to improve generalization in evaluation, both within and across 
 VO datasets.

\boldparagraph{Audio Decoder} We utilize audio labels, generally available for online videos, as an auxiliary prediction task. We note that prior work often studies such cross-modal reasoning for basic navigation scenarios~\cite{chen2020soundspaces,chen2021semantic,gan2020look} and not for in-the-wild videos where dense dynamic objects may generate significant ambient noise. In our settings, an audio signal can provide complementary information to visual information regarding the overall traffic scenario as well as ego-speed. This insight will be affirmed by our findings in Sec.~\ref{sec:analysis}, where the audio task is shown to provide synergistic supervision, both for the main VO task and when combined with other auxiliary tasks. For instance, Fig.~\ref{fig:audio} depicts how an idling ego-vehicle may generate lower audio levels, which, in conjunction with the visual scene features, can help disambiguate ego-motion from surrounding motion when stopped at intersections. As drift due to surrounding motion is a common failure mode for VO models, we further incorporate a segmentation task for dynamic objects below. 


Our audio decoder is based on a 1D U-Net architecture~\cite{schneider2023mo}, consisting of a residual 1D convolutional block~\cite{he2016deep} and an attention block~\cite{vaswani2017attention}, and reconstructs the dual-channel raw audio of the two input frames using the encoder features. We employ a two-term MSE and spectral loss, 
\begin{equation}
    \cL_{audio} = \|{\bA}_{i} - \hat{\bA}_{i}\|_2^2 + \|\text{FT}({\bA}_{i}) - \text{FT}(\hat{\bA}_{i})\|_2^2
\end{equation}
where FT represents the short-time Fourier transform~\cite{dhariwal2020jukebox}.

\boldparagraph{Segmenting Dynamic Objects} The relative motion caused by dynamic objects can often lead to inaccurate pose predictions, \eg, when stopped at a traffic light with oncoming traffic. To facilitate disambiguating potentially dynamic objects from the static background, we incorporate a segmentation prediction for pedestrian and vehicle classes~\cite{behl2020label}. As this task involves extensive manual annotation, we intentionally do not assume it is provided as part of the originally labeled dataset $\cD_L$ and instead leverage an off-the-shelf model based on  Mask R-CNN~\cite{He_2017_ICCV}. The model is pre-trained on the COCO dataset~\cite{lin2014microsoft}. We use the detector to construct a pseudo-label semantic segmentation $\tilde{\bS} \in \cR^{W \times H \times 2}$ of foreground and background in the two input frames. We leverage an FCN~\cite{long2015fully} decoder, consisting of $11$ transposed convolutional layers followed by a convolutional layer and a final sigmoid activation function, and minimize a Dice loss,
\begin{equation}
     \cL_{seg}  = 1-2\frac{\sum_{j,k,c} \tilde{\bS}_i \circ \hat{\bS}_i}{\sum_{j,k,c}  \tilde{\bS}_i^2 + \sum_{j,k,c} \hat{\bS}_i^2}
\end{equation}
where $\hat{\bS}_i \in \mathbb{R}^{H\times W \times 2}$ is the decoder predicted segmentation, and $j\in[1,H], k\in[1,W], c\in[1,2]$. As dynamic objects often cause ego-motion estimation drift, the prediction task can regularize self-training by providing a useful invariant prior (\ie, across datasets and settings) of background and foreground knowledge. Moreover, the segmentation task complements the audio task in many scenarios as dynamic objects may also generate ambient audio. 


\boldparagraph{Depth and Flow Tasks}
We explore two additional auxiliary tasks based on depth and optical flow estimation, as they potentially offer valuable information about the structure of the surroundings and the camera motion and are frequently employed in VO tasks~\cite{muller2017flowdometry, pandey2021leveraging, loo2019cnn, wang2021tartanvo}. 
We utilize an MSE as the loss function for both depth and flow tasks,
\begin{equation}
\begin{split}
    & \cL_{flow} = \|{\tilde\bF}_{i} - \hat{\bF}_{i}\|_2^2 
    \\
    & \cL_{depth} = \|{\tilde\bD}_{i} - \hat{\bD}_{i}\|_2^2
\end{split}
\end{equation}
To simplify the model, we maintain the identical decoder structure used as in the dynamic object segmentation task (see Fig.~\ref{fig:modelstructure}), with the exception of eliminating the final Sigmoid layer.




\subsection{Implementation Details}
\label{subsec:details}
Our models are trained using three NVIDIA RTX 3090 GPUs using a batch size of six.
The learning rate is set to $0.001$ and with decay $0.99$. Given the main VO objective, we set $\lambda_\theta=1$ and $\lambda_u=0.1$. Remaining auxiliary loss hyper-parameters, \ie, $\lambda_a, \lambda_s, \lambda_f, \lambda_d$, are set to $0.01$. For our semi-supervised training, we obtain a diverse set of {59,000} unlabeled samples across different geographical locations, times of day, and environmental conditions. We split the nuScenes benchmark~\cite{caesar2020nuscenes} into training, validation, and evaluation sets, to train an initial teacher model for 15 epochs. The student model is trained for 15 epochs on a mix of labeled nuScenes and pseudo-labeled YouTube data. We note that we do not employ careful ratio optimization~\cite{caine2021pseudo} when mixing the datasets without and instead solely rely on the uncertainty-based selection mechanism. We leverage data augmentation strategies, including random cropping and resizing, for improving generalization and simulating varying camera intrinsics~\cite{wang2021tartanvo}.
During inference, the model runs at $77$ FPS on a single NVIDIA GTX 3090 GPU. Additional details regarding the training and experiments can be found in the supplementary.

%% file: sec_analysis.tex
\section{Experiments}
\label{sec:analysis}

In this section, we comprehensively analyze our XVO framework. As our goal is to build generalized VO systems, we emphasize generalization ability across different datasets with various camera setups, specifically in the context of varying autonomous driving settings.

\begin{table*}[!t]
  \caption{\textbf{Analysis on the KITTI Benchmark.} We abbreviate `intrinsics-free' as {I} (\ie, a method which does not assume the intrinsics) and `real-world scale' as {S} (\ie, a method is able to recover real-world scale). To ensure meaningful comparison, we categorize models based on supervision type. Firstly, we present unsupervised learning methods, followed by supervised learning methods, then generalized VO methods, and finally our XVO ablation. In the case of TartanVO, we analyze robustness to noise applied to the intrinsics. We train two teacher models: one based on KITTI (as shown in supervised learning approaches) and the other on nuScenes (as displayed at the end of the Table with ablations). 
  }
  \label{tab:leaderboard}
  \vspace{-0.2cm}
  \centering
  \normalsize
  \resizebox{\textwidth}{!}{%
  \begin{tabular}{p{3.7cm} p{1pt} |cc|cc|cc|cc|cc|cc|cc|cc}
    \toprule
    \multirow{2}{*}{\textbf{Method}} & \multicolumn{1}{p{0.1cm}|}{}
    & \multicolumn{1}{p{0.4cm}}{\multirow{2}{*}{\centering {\,\,I}}}
    & \multicolumn{1}{p{0.4cm}|}{\multirow{2}{*}{{\,\,\,\,S}}}
    & \multicolumn{2}{c}{\centering \textbf{Seq 03}}
    & \multicolumn{2}{c}{\centering \textbf{Seq 04}}
    & \multicolumn{2}{c}{\centering \textbf{Seq 05}}
    & \multicolumn{2}{c}{\centering \textbf{Seq 06}}
    & \multicolumn{2}{c}{\centering \textbf{Seq 07}}
    & \multicolumn{2}{c}{\centering \textbf{Seq 10}} 
    & \multicolumn{2}{c}{\centering \textbf{Avg}}\\
    \cline{5-18} 
        &\multicolumn{1}{p{0.1cm}|}{}
        &\multicolumn{2}{p{1.5cm}|}{}
        & $t_{rel}$ &$r_{rel}$ 
        & $t_{rel}$ &$r_{rel}$ 
        & $t_{rel}$ &$r_{rel}$ 
        & $t_{rel}$ &$r_{rel}$ 
        & $t_{rel}$ &$r_{rel}$ 
        & $t_{rel}$ &$r_{rel}$ 
        & $t_{rel}$ &$r_{rel}$ \\
        \bottomrule
    \multicolumn{14}{l}{\textit{Unsupervised Methods:}} \\
    \toprule
        

    SfMLearner~\cite{zhou2017unsupervised}    & & \xmark & \xmark & 10.78 & 3.92 &  4.49 & 5.24 &  18.67 & 4.10 &  25.88 & 4.80 &  21.33 &6.65 &  14.33 & 3.30 & 15.91 & 4.67 \\

    GeoNet~\cite{yin2018geonet}    & & \xmark & \xmark & 19.21 & 9.78  & 9.09 & 7.55 &  20.12 & 7.67 &  9.28 & 4.34 &  8.27  & 5.93  & 20.73 & 9.10 & 14.45 & 7.40 \\

    Zhan~\textit{et al.}~\cite{zhan2018unsupervised}    & & \xmark & \cmark & 15.76 & 10.62  & 3.14 & 2.02  & 4.94 & 2.34  & 5.80 & 2.06  & 6.49 &3.56  & 12.82 & 3.40 & 8.16 & 4.00 \\

    UnDeepVO~\cite{li2018undeepvo}    & & \xmark & \cmark & 5.00 & 6.17 & 4.49 & 2.13 & 3.40 & 1.50 & 6.20 & 1.98 & 3.15 & 2.48  & 10.63 & 4.65 & 5.48 & 3.15 \\




    \bottomrule
    \multicolumn{14}{l}{\textit{Supervised Methods:}} \\
    \toprule

    DeepVO~\cite{wang2017deepvo}   & & \cmark & \cmark & 8.49 & 6.89 & 7.19 & 4.97  & 2.62 & 3.61  & 5.42 & 5.82  & 3.91 &4.60  & 8.11 & 8.83 & 5.96 & 5.79 \\

    ESP-VO~\cite{wang2018end}   & & \cmark & \cmark & 6.72 & 6.46 &  6.33 & 6.08 &  3.35 & 4.93 &  7.24 & 7.29 &  3.52 & 5.02 &  9.77 & 10.2 & 6.16 & 6.66\\

    GFS-VO~\cite{xue2019guided}    & & \cmark & \cmark & 5.44 & 3.32  & 2.91 & 1.30 & 3.27 & 1.62 & 8.50 & 2.74 & 3.37 & 2.25 & 6.32 & 2.33 & 4.97 & 2.26 \\

    Xue~\etal~\cite{xue2019beyond} & & \cmark & \cmark & \textbf{3.32} & 2.10 & 2.96 & 1.76 & 2.59 & 1.25 & 4.93 & 1.90 & \textbf{3.07} & \textbf{1.76}  & 3.94 & 1.72 & 3.47 & \textbf{1.75}\\

    \textbf{Our Teacher (KITTI)} & & \cmark & \cmark & ${3.46}$ & $\mathbf{2.00}$ & $\mathbf{1.67}$ & $\mathbf{0.70}$ & $\mathbf{2.12}$ & $\mathbf{0.92}$ & $\mathbf{3.92}$ & $\mathbf{1.46}$ & ${5.93}$ & ${3.96}$ & $\mathbf{3.31}$ & $\mathbf{1.52}$ & $\mathbf{3.40}$ & ${1.76}$\\
    
    
    \bottomrule
    \multicolumn{14}{l}{\textit{Baseline Generalized VO Methods:}} \\
    \toprule
    TartanVO~\cite{wang2021tartanvo}    & & \xmark & \xmark & 4.20 & 2.80  & 6.19 & 4.35  & 5.84 & 3.24  & 4.21 & 2.51  & 7.11 & 4.96  & 8.00 & 3.21 & 5.93 & 3.51\\

    
    TartanVO (10\% Noise)    & & \xmark & \xmark & 9.33 & 3.12  & 10.88 & 4.71  & 11.77 & 5.39  & 11.88 & 4.52  & 14.70 &10.74  & 11.76 & 3.61 &11.72 &5.35 \\
    
    TartanVO (20\% Noise)    & & \xmark & \xmark & 17.79 & 4.42 & 21.58 & 5.04  & 20.12 & 8.54  & 18.80 & 6.26  & 21.34 &16.27  & 17.45 & 5.03 & 19.51 & 7.59 \\
    
    TartanVO (30\% Noise)    & & \xmark & \xmark & 25.89 & 7.06  & 34.91 & 4.54  & 22.48 & 10.17  & 19.32 & 5.23  & 19.40 &13.33  & 25.06 & 8.43 &24.51 & 8.13\\


    \bottomrule
    \multicolumn{14}{l}{\textit{Proposed Generalized VO Methods:}} \\
    \toprule 
    

    Our Teacher (nuScenes) & & \cmark & \cmark & 26.78 & 4.92 & 26.02 & 2.42 & 23.65 & 8.85 & 23.97 & 6.47 & 30.66 &20.32 & 20.57 & 6.01 &25.27 &8.17\\

    Student w/o Filter & & \cmark & \cmark & 26.98 & 9.68  & 22.56 & 2.15  & 14.77 & 5.83 & \textbf{11.38} & \textbf{1.62} & 16.45 &9.35 & 20.23 & 8.99 &18.73 & 6.27\\

    Student & & \cmark & \cmark  & 20.30 & 3.97 & 16.33 & 1.57  & 11.12 & 4.19 & 15.60 & 5.69  & 7.77 & 3.48  & 19.91 & 5.59 &15.17 &4.08 \\

    \textbf{XVO} & & \cmark & \cmark  & \textbf{14.53} & \textbf{3.93} & \textbf{16.29} & \textbf{0.96} & \textbf{8.31} & \textbf{2.76} & 15.31 & 5.49  & \textbf{5.86} &\textbf{3.00} & \textbf{12.17} & \textbf{3.45} & \textbf{12.08} &\textbf{3.27}\\


    
    \bottomrule
    
  \end{tabular}%
  \vspace{-0.1in}
}
\end{table*}

\begin{table}[!ht]
\scriptsize
\setlength\tabcolsep{2pt}
\renewcommand\arraystretch{1.1}
\centering
\caption{\textbf{Average Quantitative Results across Datasets.} We test on KITTI (sequences 00-10), Argoverse 2, and the unseen regions in nuScenes. All results are the average over all scenes. We present translation error, rotation error and scale error. Approaches such as TartanVO do not estimate real-world scale but may be aligned with ground truth (GT) scale in evaluation. A, S, F, D are the abbreviation of Audio, Seg, Flow, Depth.} 
\label{tab:nuscenes}
{
\begin{tabular}{p{2.5cm}|ccc|ccc|ccc}
\toprule
\multirow{2}{*}{\textbf{Method}} & 
\multicolumn{3}{c|}{\textbf{KITTI 00-10}} &
\multicolumn{3}{c|}{\textbf{Argoverse 2}} &
\multicolumn{3}{c}{\textbf{nuScenes}}
\\  \cline{2-10} 
        & $t_{err}$ &$r_{err}$ & $se$ 
        & $t_{err}$ &$r_{err}$ & $se$ 
        & $t_{err}$ &$r_{err}$ & $se$ \\ 

    \bottomrule
    \multicolumn{9}{l}{\textit{Baseline Generalized VO Methods:}} \\
    \toprule 
    
    
    {TartanVO w/ GT Align} & 6.37 & 3.32 & / & 8.55 & 5.77 & / & 9.61 & 6.83 & / \\
    
    {TartanVO w/o GT Align} &  21.67 & 3.33 &0.29  & 41.11 & 5.77 & 0.40 & 28.23 & 6.83 & 0.29 \\

    \bottomrule
    \multicolumn{9}{l}{\textit{Proposed Generalized VO Methods:}} \\
    \toprule 
    
    {Teacher (nuScenes)} & 26.16 & 6.84 & 0.25 & 10.89 & 3.40 & 0.16 &15.93 & 6.73 & 0.20 \\
    
    {Student w/o Filter} & 20.64 & 5.68 & 0.21 & 10.80 & 7.33 & 0.14 & 9.32 & 4.60 & 0.14 \\
    
    {Student} & 17.04 & 4.02 & 0.16 & 9.16 & 3.40 & 0.14 & 10.54 & 3.94 & 0.13 \\
    
    
    {Student+Seg}  & 16.31 & 3.77 & 0.16 & 9.17 & \textbf{3.18} & 0.13 & 11.35 & 4.05 & 0.14 \\
    

    {Student+Flow} &15.60&3.19&0.19 &9.04&4.45&0.13 &\textbf{9.13}&4.06&\textbf{0.13} \\

    {Student+Depth} &17.49&3.89&0.20 &9.25&4.11&0.13 &11.86&6.46&0.15 \\
    
     
    
    {Student+Audio} & \textbf{14.37} & \textbf{3.06} & \textbf{0.16}  & \textbf{8.00} & \textbf{3.08} & \textbf{0.12} & \textbf{9.26} & \textbf{3.20} & \textbf{0.12} \\

    
    {Student+Audio+Seg} & \textbf{14.20} & \textbf{3.02} &\textbf{0.16} & 8.67 & 3.63 & 0.13 & 11.29 & 3.70 & 0.14 \\

    {Student+S+F+D} &18.23&3.88&0.21 &8.79&4.89&0.13 &8.93&\textbf{3.44}&0.13 \\

    {Student+A+S+F+D} &16.74&4.40&0.18 &\textbf{7.89}&3.54&\textbf{0.12} &9.98&4.36&0.15 \\

    
    

    \bottomrule
\end{tabular}}
\end{table}

\boldparagraph{Datasets} To understand the role of cross-modal self-training on model generalization, we evaluate our proposed XVO method using three commonly employed datasets, KITTI~\cite{Geiger2012CVPR}, nuScenes~\cite{caesar2020nuscenes} and Argoverse 2~\cite{wilson2021argoverse}. Out of the three, KITTI is the most popular VO benchmark, consisting of $11$ sequences $00$-$10$ with ground truth. As KITTI is an older benchmark (2012), its camera intrinsics vary significantly from the other two benchmarks. nuScenes consists of about 15 hours of driving data (totaling $197$,$000$ images) from four regions in Boston and Singapore: Boston-Seaport, Singapore-OneNorth, Singapore-Queenstown, and Singapore-HollandVillage. In contrast to KITTI which was captured in sunny driving with mostly static objects, nuScenes incorporates complex real-world driving maneuvers in dense streets and various conditions, \eg, nighttime, difficult illumination conditions with low visibility, as well as artifacts on the camera lens, such as rain droplets or dirt. Finally, Argoverse 2 is a large dataset with $1$,$000$ driving sequences across six US cities. We leverage a test dataset that includes 150 sequences and $48$,$022$ images. 

\boldparagraph{Procedure and Baselines} We generally train within one region on nuScenes (HollandVillage) and evaluate the remaining regions and datasets. This is in contrast to prior evaluation procedures where models can learn to memorize the scale and camera setup without generalization through training and testing on the same camera setup and similar environments. We also directly compare with prior state-of-the-art using the standard KITTI protocol~\cite{li2018undeepvo,yin2018geonet}. As our approach does not leverage known intrinsics, we separate approaches that do assume such knowledge in their pipeline to ensure meaningful analysis. We further indicate whether methods predict pose with absolute scale, as some methods output up-to-scale estimates and use the ground-truth scale to align and evaluate their model, \eg, TartanVO~\cite{wang2021tartanvo}. Nonetheless, TartanVO is one of the few approaches that have shown generalization across datasets without the need to fine-tune or perform online adaptation strategies and is therefore our main baseline. 

\boldparagraph{Metrics} We follow standard evaluation metrics of average translation $t_{rel}$ (in \%) and rotation $r_{rel}$ (in degrees per 100 meters) errors, computed over all possible subsequences within a test sequence of lengths $(100,...,800)$ meters~\cite{wang2021tartanvo,Geiger2012CVPR}. We refer the reader to the KITTI leaderboard for more details regarding the metric. However, we observe prior measures to only provide a proxy evaluation of real-world scale predictions as the errors could potentially vary along the trajectory independently of trajectory-level measures (our supplementary contains additional details). To explore the benefits of semi-supervised training on real-world scale estimation, we sought to directly quantify scale recovery within consecutive frames in a single metric. We therefore also report the \textit{average scale error ($se$)} over predicted and ground-truth translation,
\begin{equation} 
se = 1-\min\left(\frac{\|\hat{\bt}\|_2}{\max(\|\bt\|_2,\epsilon)}, \frac{\|\bt\|_2}{\max(\|\hat{\bt}\|_2,\epsilon)}\right) 
\end{equation}
where $\epsilon$ prevents dividing by zero. 

\subsection{Results}

We examine the role of the main components in the proposed framework below. Complete ablation, \eg, across modality combinations and training settings, can be found in the supplementary.  




\boldparagraph{Teacher Model Performance} Table~\ref{tab:leaderboard} compares our proposed encoder architecture for the main VO task with prior methods. When trained in a supervised learning manner on KITTI, our teacher model achieves the lowest translation error of $3.4 \%$ even without access to camera intrinsics or multi-step optimization. This suggests that basic modifications to underlying network structure, \eg, through an improved encoder and attention-based mechanism, can result in significant gains for the monocular VO task. Given the effective network structure, we now turn to analyzing the benefits of the proposed semi-supervised framework. 

\boldparagraph{Semi-Supervised VO Training}
Table~\ref{tab:leaderboard} also analyzes the generalization performance of the proposed semi-supervised learning framework on KITTI. Specifically, we show our initial teacher model that is trained on the nuScenes (HollandVillage) dataset to not generalize well to the KITTI testing set ($25.27\%$ translation and $8.17^{\circ}$ rotation error) due to domain shift and differing camera settings. However, after semi-supervised training, the errors for the student model are reduced by $40\%$ and $50\%$ in translation and rotation errors, respectively. The best self-trained model with auxiliary tasks (complete ablation can be found in the supplementary) results in further student performance gains, \eg, a further reduction in translation error by $20\%$. We also compare with the most related TartanVO~\cite{wang2021tartanvo} baseline which utilizes the ground-truth for scale alignment and has access to camera intrinsics. However, even with the ground-truth alignment, TartanVO exhibits quick degradation with minimal noise in the intrinsics (enabling a more fair comparison as our method is not provided these as input). Moreover, we explore the generalization of our training framework by evaluating on various datasets in Table~\ref{tab:nuscenes}. We emphasize that none of the trained models have access to samples from Argoverse 2 or KITTI dataset during training. By predicting real-world scale, our student model with all auxiliary tasks outperforms the baseline TartanVO in all three datasets, \eg, by $80\%$ in translation and $70\%$ scale error on Argoverse 2, without any ground-truth alignment. This indicates the proposed method to improve reasoning over scale and scene semantics across arbitrary conditions. 

\begin{figure}
    \centering
    \includegraphics[width=0.47\textwidth]{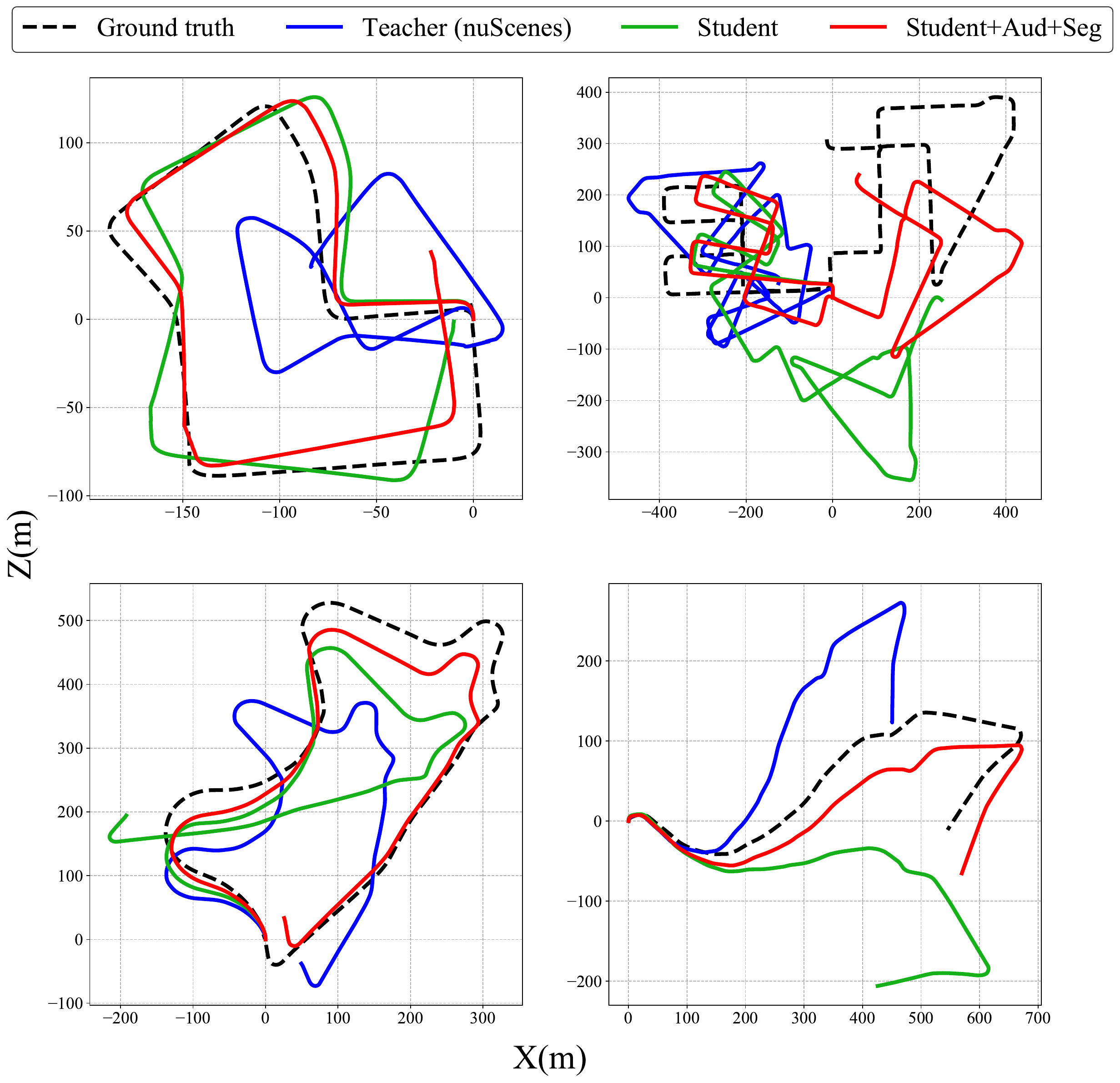}
    \caption{\textbf{Qualitative Analysis on KITTI.} We find that incorporating audio and segmentation tasks as part of the semi-supervised learning process significantly improves ego-pose estimation on KITTI.}
    \label{fig:traj}
   \vspace{-0.2cm}
\end{figure}

\vspace{-0.2cm}

\boldparagraph{Impact of Uncertainty-Aware Sample Selection} When inspecting the various pseudo-labels, we observed many cases of drift and incorrect predictions due to the harsh generalization settings. Hence, the uncertainty-aware pseudo-label selection mechanism plays a crucial role in the semi-supervised learning process. As shown in Table~\ref{tab:leaderboard} and Table~\ref{tab:nuscenes}, discarding pseudo-labels with low confidence consistently improves performance, both with and without multi-modal supervision. We notice how a student model without the uncertainty-aware sample removal (\ie, `Student w/o Filter') provides only mild improvements compared to the teacher. Once noisy samples are filtered out of the dataset, the performance on KITTI and nuScenes improve significantly, \eg, from 26.16 to 17.04 and 15.93 to 10.54 translation error respectively.

\boldparagraph{Ablation on Auxiliary Tasks} We sought to understand the role of the various explored auxiliary tasks, \ie, audio, segmentation, depth, and flow. 
We first analyze the impact of adding an audio reconstruction task for the VO problem. Although extracted audio includes some ambient noise, we can see that XVO consistently benefits from the proposed audio supervision across the evaluation datasets. This can be explained by the consistent quality of the ground-truth audio labels, \ie, when compared to the noise in pseudo-labels generated by the auxiliary prediction models on our unconstrained videos. In general, we find that audio, segmentation, and flow tasks result in better performance when compared to the depth prediction task. While prior research often leverages monocular depth prediction for improving VO on KITTI, this is a significantly challenging task in more general settings which results in noisier pseudo-labels. 
We also investigate various combinations of auxiliary branches and find the combination of segmentation and audio branch performs better than a single auxiliary task on KITTI. While this is encouraging, KITTI contains simpler scenarios with relatively few dynamic traffic participants. In such simpler settings, our segmentation branch can be used to obtain reliable pseudo-labels and learn efficient generalized features. However, this finding does not extend to nuScenes and Argoverse 2 which frequently contain dense and dynamic scenes. We also find that simply adding prediction tasks does not provide further gains due to the pseudo-label noise and a more brittle and difficult optimization process. Complete ablations on auxiliary tasks can be seen in our supplementary. 


\subsection{Qualitative Results}
Fig.~\ref{fig:traj} depicts the prediction of driving trajectories on KITTI sequences 7, 8, 9, and 10.  The trajectory predicted using the teacher model that is trained on nuScenes is not able to recover scale accurately. Due to the semi-supervised training process, the student model is shown to have better scale recovery and generalization despite the lack of calibration knowledge. Nonetheless, the student model fails to estimate accurate rotation in more challenging scenes on KITTI, \eg, top right and bottom left scenarios. Finally, the cross-modal trained model is shown to robustly estimate translation, rotation, and scale, even in the most complicated route in Fig.~\ref{fig:traj}-top right. Additional qualitative examples are provided in the supplementary.

%% file: sec_conclusion.tex
\section{Conclusion}
\label{conclusion}



In this paper, we present XVO, a novel method for generalized visual odometry estimation via cross-modal self-training. Our efficient network structure achieves state-of-the-art results on KITTI, despite having no knowledge of camera parameters or multi-frame optimization as in prior methods. Moreover, our framework leverages a mixed-label semi-supervised setting over a large dataset of internet videos to further enhance generalization performance. Specifically, we show that additional auxiliary segmentation and audio reconstruction tasks can significantly impact cross-dataset generalization. Our trained VO models can be used across platforms and settings without fine-tuning, \ie, due to general reasoning over semantic visual characteristics of scenes. Moreover, our training settings of improving the performance of a model that is initially trained on a small and restricted dataset are broadly applicable to various robotics use-cases. We hope our work can inspire future researchers to explore scalable VO models that can benefit a broad range of applications. Given the limited utility of combining multiple auxiliary tasks in our settings, a future direction would be to study better methods for learning from noisy and diverse auxiliary pseudo-labels. Moreover, while we achieved state-of-the-art results with a two-frame approach, multi-frame optimization could provide further benefits by alleviating drift.    


\boldparagraph{Acknowledgments}
We thank the Red Hat Collaboratory and National Science Foundation (IIS-2152077) for supporting this research.





%% file: sec_supp.tex
\begin{abstract}
In this supplementary document, we first discuss additional implementation details (Sec.~\ref{sec:impl}), including our network architecture, training settings, and employed evaluation metrics. Next, we provide ablative analysis across different baselines, datasets, and auxiliary tasks (Sec.~\ref{sec:ablate}). Finally, we include additional qualitative examples (Sec.~\ref{sec:qual}, also shown in our supplementary video).

\end{abstract}


\section{Implementation Details}
\label{sec:impl}

\subsection{Network Architecture} 
As mentioned in the main paper, we learn a unified feature extractor based on a MaskFlownet encoder~\cite{zhao2020maskflownet} that is followed by four self-attention layers~\cite{dosovitskiy2020image,chen2021transunet}.
The encoder structure for the initial teacher model and the subsequent cross-modal student model is kept the same, and provides state-of-the-art results on KITTI. Task decoders are then added over the shared visual encoder and supervised by auxiliary audio, visual, and motion prediction tasks in order to  encourage the model to extract information that is relevant across a wide range of conditions, including potentially unseen environments.  

\boldparagraph{Task Decoders} Due to their similar output representations, we incorporate segmentation, depth, flow auxiliary tasks using a similar decoder structure (also outlined in the main paper). The goal of these three auxiliary tasks is to encourage the visual encoder to identify geometric and motion cues that can also be relevant for the primary VO task. The segmentation decoder employs an FCN~\cite{long2015fully} decoder, consisting of $11$ transposed convolutional layers followed by a convolutional layer and a final sigmoid activation function. The output size is $289\times296$. The optical flow decoder shares the same structure but without a sigmoid and pretrained on the Flying Chairs~\cite{dosovitskiy2015flownet} dataset. The depth decoder was pretrained using NYU Depth V2~\cite{Silbermaneccv2012}.

\subsection{Data}

\boldparagraph{Augmentation Strategies}
To generalize model performance, we use several data augmentation techniques including random cropping and resizing which simulates varying camera intrinsics.

\boldparagraph{YouTube Videos} YouTube has ample navigation videos, often with corresponding audio. We searched for high-resolution driving dash camera videos and downloaded diverse data with different times of day (both daylight and nighttime), weathers (rain, snow, sun), environments (urban, rural, suburban), and location (Boston, Washington, London, Singapore, Paris, Switzerland, Milan, Ireland, Tokyo). The original video data is about 100 minutes at 30FPS, which we then subsample by three to 10FPS to obtain the final dataset.

\subsection{Training Details}
All models are trained for 15 epochs using Stochastic Gradient Descent with a batch size of six and input image size of $640\times384$~\cite{ruder2016overview, shangguan2021neural}. Training takes approximately two days. We perform extensive ablation on the role of the various auxiliary tasks for model training in Sec.~\ref{sec:ablate}. 
We empirically set the entropy threshold in Eqn. 5 of the main paper as $\tau_u=-5.668$. As we do not find it beneficial to iterate over semi-supervised training (\ie, with the new student as the teacher), we only perform one iteration of teacher-student training.


\subsection{Evaluation Metrics}

We sought to accurately measure the ability of our model to predict general real-world pose, including absolute scale, throughout a trajectory. In this section, we define our metrics, including a proposed scale error metric which can more accurately account for scale errors in prediction along a trajectory. While our best performing model is shown to outperform other models in terms of all three defined metrics, we demonstrate the semi-supervised learning process to consistently improve scale estimation as well (discussed further in Sec.~\ref{sec:ablate}). 


\begin{figure}
    \centering
    \includegraphics[width=1.9in]{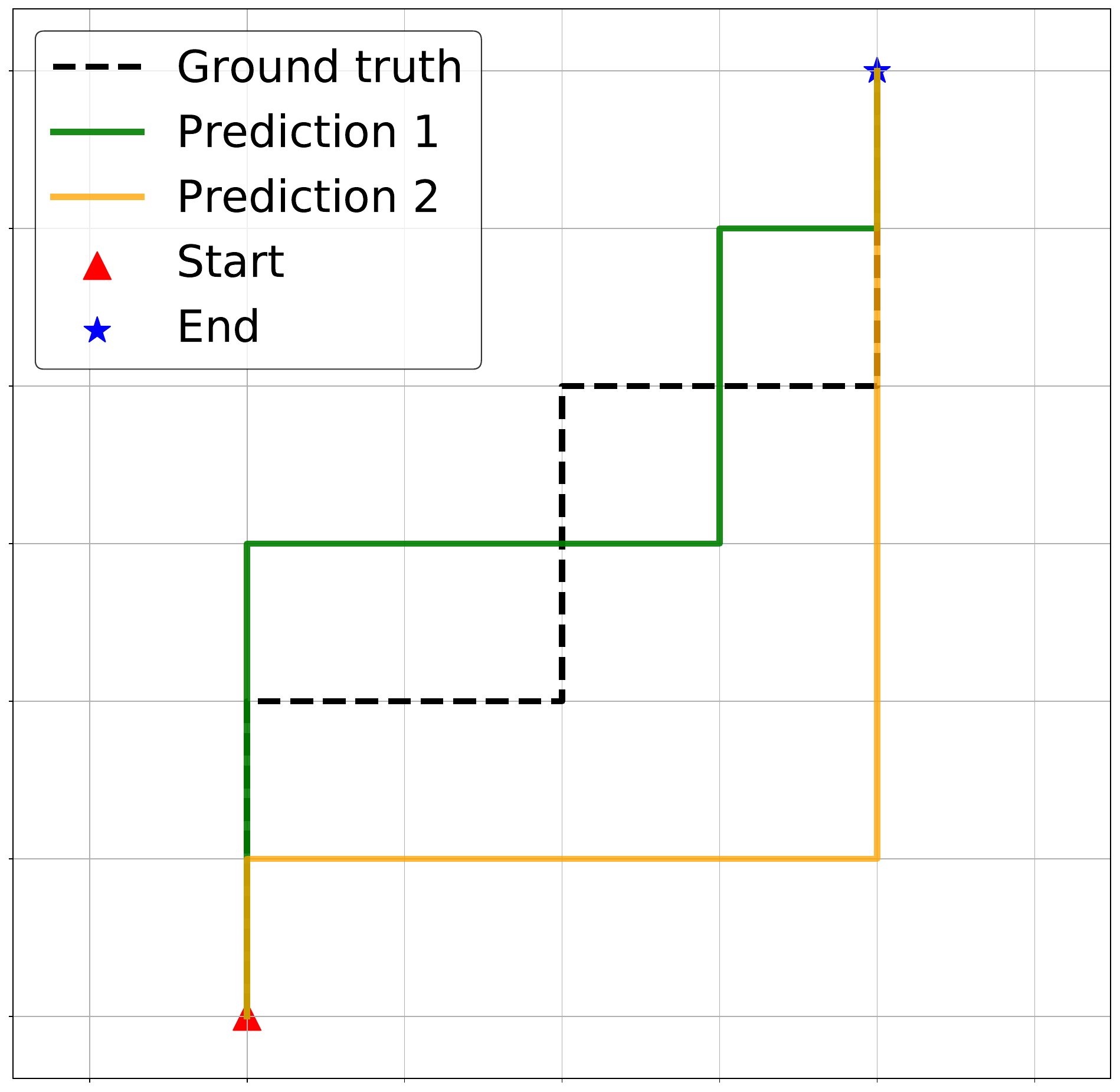}
    \caption{\textbf{Low Translation and Rotation Error Despite Inaccurate Predictions.} Given our emphasis on estimating real-world pose, including absolute scale, across diverse platforms, we highlight issues with the definition of standard translation and rotation errors. As standard VO metrics are computed over trajectory end-points, the trajectory itself can vary arbitrarily between the end-points while maintaining low error. In the example shown, all predicted trajectories will have zero translation and rotation error. However, using an average scale error will demonstrate a difference. The scale error for Prediction $1$ and Prediction $2$ is $0.33$ and $0.30$ respectively. Therefore, introducing the scale error provides a more holistic evaluation of model performance while directly measuring scale. }
    \label{fig:explain}
\end{figure}

\boldparagraph{Translation and Rotation Errors at Trajectory End-Points} Standard visual odometry metrics average relative pose errors at fixed distances along the trajectory~\cite{Geiger2012CVPR,kummerle2009measuring,prokhorov2019measuring}. 
We follow common metrics to compute translation and rotation errors which are computed over all possible trajectory subsequences of length $(100, \dots, 800)$ meters. Given a trajectory of length $l$, standard error metrics measure relative pose between an estimated pose and the ground-truth pose of the last frame with respect to the first frame, $[\bR^{\prime}|\bt^{\prime}]$. Translation and rotation errors are computed as averaged translational (\%) drift and averaged rotational drift (${^\circ}{/}100m$):
\begin{equation} \label{eq:trans}
t_{rel} = \|\bt^\prime\|_2*\frac{100}{l}
\end{equation}
\begin{equation} \label{eq:rot}
r_{rel} = \arccos(\max\left(\min\left(d,1\right), -1\right))*\frac{180}{\pi}*\frac{100}{l}
\end{equation}
where $d=\frac{tr(\bR^\prime)-1}{2}$.

\boldparagraph{Issues with Standard Metrics}
Relative translation and rotation errors can measure an error drift over the trajectory subsequence. However, these measures can neglect instantaneous errors along the trajectory subsequence. Here, multiple differing trajectories can all have similar rotation and translation errors, given similar translation and rotation at their end-points. Critically, the \textit{translation error can be low even if the estimated trajectory itself is far from the ground-truth trajectory} prior to the end-point used for evaluation. To emphasize, as long as the estimated translation and rotation at the end of the trajectory is same with ground-truth, translation error and rotation error can be low despite potentially high two-frame inaccuracies. 


We present a toy example in Fig. \ref{fig:explain} and a trajectory of six consecutive time steps with ground-truth relative rotations of $[0^{\circ},90^{\circ},0^{\circ}]$, $[0^{\circ},-90^{\circ},0^{\circ}]$, $[0^{\circ},90^{\circ},0^{\circ}]$, $[0^{\circ},-90^{\circ},0^{\circ}]$, $[0^{\circ},0^{\circ},0^{\circ}]$, and ground-truth relative translations is kept fixed at each time step $[0,0,20]$. In Prediction $1$, assume our predicted relative rotations are similar to the ground-truth and our predicted relative translation are $[0,0,30]$,$[0,0,30]$,$[0,0,20]$,$[0,0,10]$,$[0,0,10]$, respectively. In Prediction $2$, assume our predicted relative rotations are $[0^{\circ},90^{\circ},0^{\circ}]$, $[0^{\circ},-90^{\circ},0^{\circ}]$, $[0^{\circ},0^{\circ},0^{\circ}]$, $[0^{\circ},0^{\circ},0^{\circ}]$, $[0^{\circ},0^{\circ},0^{\circ}]$ and our predicted relative translations are $[0,0,10]$,$[0,0,40]$,$[0,0,10]$,$[0,0,20]$,$[0,0,20]$, respectively. Once we recover the trajectories based on the ground-truth relative pose and predicted relative pose, all trajectories share the same endpoint with same direction. Here, while the model fails at two-frame prediction, the translation error and rotation error are both zero. Although this example may seem contrived, we sought a more accurate evaluation of our model which directly estimates real-world scale using two-frame input, as discussed next.

\boldparagraph{Scale Error} 
Towards a more accurate evaluation of real-world scale along the trajectory, we also compute and average the two-frame scale error (defined in the Eqn. 10 in the main paper). We therefore also report the \textit{average scale error ($se$)}, which is an error computed over predicted and ground-truth translation,
\begin{equation} 
se = 1-\min\left(\frac{\|\hat{\bt}\|_2}{\max(\|\bt\|_2,\epsilon)}, \frac{\|\bt\|_2}{\max(\|\hat{\bt}\|_2,\epsilon)}\right).
\end{equation}
Here, the scale error is always computed over two sequential frames, as opposed to over end-points of a multi-step trajectory, and therefore provides a more holistic evaluation of model performance while directly measuring scale. The scale error of Prediction $1$ and Prediction $2$ in Fig. \ref{fig:explain} are then $0.33$ and $0.30$, respectively.

\section{Ablation Studies}
\label{sec:ablate}

This section provides comprehensive analysis with various modalities and training settings using our cross-dataset evaluation. In Sec.~\ref{subsec:tvo} we discuss comparison with our baseline, TartanVO~\cite{wang2021tartanvo}. Next, we explore the impact of various auxiliary prediction tasks in Sec.~\ref{subsec:mod} as well as the uncertainty-based filtering process for the pseudo-labels in Sec.~\ref{subsec:filter_supp} and the impact of additional semi-supervised learning iterations (Sec.~\ref{subsec:iter}). 



\subsection{TartanVO Baseline}
\label{subsec:tvo}
Our main comparative baseline is TartanVO~\cite{wang2021tartanvo} as it is one of the few VO methods that can perform cross-dataset generalization without fine-tuning. Moreover, TartanVO does not leverage long-term memory to correct for drift (\ie, the input is two-frame, as with our model). However, TartanVO \textit{leverages the intrinsic parameters} to adapt across camera settings as input to the network, while our method does not. Moreover, in evaluation, the approach aligns the predicted pose to the ground-truth, \ie, by only estimating pose up to a scale factor which is determined using the ground-truth in evaluation. In contrast, our study emphasizes accurate real-world scale prediction. Since up-to-scale prediction can limit the applicability of the method in practice, we analyze performance with and without the ground-truth scale alignment step in Table~\ref{tab:across}. As shown in the table, the removal of the ground-truth alignment step allows us to better evaluate and compare with our models. On average across the dataset, our top performing models (\eg, Student+Audio) outperform TartanVO both with and without ground-truth alignment by (\eg, 10.54 vs. 14.84 and 3.11 vs. 5.31 for the translation and rotation errors, respectively). We note that we cannot compute a scale error for methods that estimate pose up to a scale.

\subsection{Modality Ablation}
\label{subsec:mod}

Table~\ref{tab:across} shows the result of training a teacher model on nuScenes and evaluating on the three benchmarks, including unseen portions of nuScenes. We then train various student models with different auxiliary tasks. For legibility, we also color the top three performing methods within each dataset and metric, as well as summarized performance across all settings (however, while KITTI only has 11 sequences, nuScenes has 765 testing sequences). We also note that our results are with \textit{pseudo labels of depth and motion cues and not ground-truth} which is often employed by related studies in cross-task~\cite{zamir2020robust,zamir2018taskonomy}. In our semi-supervised settings, the pseudo-labels can still contain noise despite our filtering mechanism. 

\boldparagraph{Results on KITTI}  On KITTI, we find adding depth and audio-based models (both audio, audio+segmentation, and audio+segmentation+flow) to perform best. This finding is consistent both with the overall averages and the per-sequence results for KITTI (shown in Table~\ref{tab:leaderboard_supp}). As depth provides a general cue regarding scene geometry, we find it to benefit pose estimation performance on KITTI (translation error with the student model drops from 17.04 to 13.09), but to a lesser extent on the other datasets where the audio-based approaches perform best. This can be explained by inspecting the KITTI scenes, which contain simpler layouts with few dynamic objects compared to Argoverse and nuScenes (where flow and segmentation-based prediction is also shown to be helpful). In such scenarios, depth cues can be more complex in contrast with the others, \eg, audio or motion cues.

\begin{table*}[!t]
\scriptsize
\setlength\tabcolsep{2pt}
\renewcommand\arraystretch{1.1}
\centering
\caption{\textbf{Model Ablation.} We train models on nuScenes and test our approach using various model settings and datasets (KITTI sequences 00-10, Argoverse 2, and unseen areas in nuScenes). We also show overall average results. TartanVO~\cite{wang2021tartanvo}, our main baseline, leverages the intrinsics and does not predict absolute scale directly, but instead evaluates with scale alignment using the ground-truth scale at each time step. Results are shown for translation, rotation, and scale errors. `w/o Filter' refers to removal of the proposed uncertainty-aware pseudo-label sample selection mechanism. Lowest three errors are highlighted within each column (when two numbers are identical to two decimal places, we compare their original precision). 
 } 
\label{tab:across}
\begin{tabular}{p{3.5cm}|ccc|ccc|ccc|ccc}
\toprule
\multirow{2}{*}{\textbf{Method}} & 
\multicolumn{3}{c|}{\textbf{KITTI 00-10}} &
\multicolumn{3}{c|}{\textbf{Argoverse 2}} &
\multicolumn{3}{c|}{\textbf{nuScenes}}
&
\multicolumn{3}{c}{\textbf{Average}}
\\  \cline{2-13} 
        & $t_{err}$ &$r_{err}$ & $se$ 
        & $t_{err}$ &$r_{err}$ & $se$
        & $t_{err}$ &$r_{err}$ & $se$
        & $t_{err}$ &$r_{err}$ & $se$ \\ \hline
    
    
    {TartanVO w/ GT Alignment~\cite{wang2021tartanvo}} & 6.37 & 3.32 & / & 8.55 & 5.77 & / & 9.61 & 6.83 & / & 8.17  & 5.31 & / \\
    
    {TartanVO w/o GT Alignment} &  21.67 & 3.33 &0.29  & 41.11 & 5.77 & 0.40 & 28.23 & 6.83 & 0.29  & 30.34 & 5.31 & 0.33\\
    \hline\hline 
    
    {Teacher (nuScenes)} & 26.16 & 6.84 & 0.25 & 10.89 & 3.40 & 0.16 &15.93 & 6.73 & 0.20  & 17.66 & 5.66 & 0.20 \\
    \hline 
    
    {Student w/o Filter} & 20.64 & 5.68 & 0.21 & 10.80 & 7.33 & 0.14 & 9.32 & 4.60 & 0.14 & 13.59 & 5.87 & 0.16 \\
    
    {Student} & 17.04 & 4.02 & 0.16 & 9.16 & 3.40 & 0.14 & 10.54 & 3.94 & 0.13 & 12.24 & 3.79 & 0.14 \\
    
    {Student+Seg w/o Filter} & 16.65 & 3.42 & 0.20 & 9.83 & 4.63 & 0.13 & \cellcolor{lb}8.33 & 3.75 & 0.12  & 11.60 & 3.93 & 0.15\\
    
    {Student+Seg}  & 16.31 & 3.77 & 0.16 & 9.17 & 3.18 & 0.13 & 11.35 & 4.05 & 0.14 & 12.28 & 3.67 & 0.14 \\

    {Student+Flow w/o Filter} &20.47&5.65&0.22 &10.74&5.95&0.13 &10.02&4.87&0.13 & 13.74 & 5.49 & 0.16 \\
    {Student+Flow} &15.60&3.19&0.19 &9.04&4.45&0.13 &9.13&4.06&0.13  & \cellcolor{lb}11.26 & 3.90 & 0.15\\
    {Student+Depth w/o Filter} &19.15&4.71&0.18 &8.64&3.70&0.13 &11.48&3.80&0.14  & 13.09 & 4.07 & 0.15\\
    {Student+Depth} &\cellcolor{lb}13.09&\cellcolor{lo}3.03&0.16 &9.91&\cellcolor{lo}3.08&0.14 &10.86&5.55&0.14  & 11.29 & 3.89 & 0.15\\
     
    {Student+Audio w/o Filter} & 16.45 & 3.88 & 0.19 & 9.83 & 5.38 & 0.14 & 9.02 & 4.03 & 0.14 & 11.77 & 4.43 & 0.16\\
    {Student+Audio} & \cellcolor{lb}14.37 & 3.06 & \cellcolor{lm}0.16  & 8.00 & \cellcolor{lo}3.08 & \cellcolor{lm}0.12 & 9.26 & \cellcolor{lo}3.20 & \cellcolor{lm}0.12  & \cellcolor{lb}10.54 & \cellcolor{lo}3.11 & \cellcolor{lm}0.13\\
    
    
    {Student+Audio+Seg} & \cellcolor{lb}14.20 & \cellcolor{lo}3.02 &\cellcolor{lm}0.16 & 8.67 & 3.63 & 0.13 & 11.29 & 3.70 & 0.14  & 11.39 & 3.45 & \cellcolor{lm}0.14\\
    Student+Audio+Depth &15.45&3.86&0.18 &8.43&4.51&0.13 &10.44&4.94&0.14  & 11.44 & 4.44 & 0.15\\
    Student+Seg+Depth &15.07&3.98&\cellcolor{lm}0.15 &10.03&\cellcolor{lo}3.10&0.15 &9.58&3.40&0.13  & 11.56 & 3.49 & 0.14\\
    {Student+Audio+Flow} &15.84&3.09&0.19 &\cellcolor{lb}7.76&3.59&\cellcolor{lm}0.12 &\cellcolor{lb}8.44&3.45&\cellcolor{lm}0.12 & \cellcolor{lb}10.68 & \cellcolor{lo}3.38 & \cellcolor{lm}0.14\\
    {Student+Flow+Depth} &17.08&3.07&0.20 &\cellcolor{lb}7.84&3.85&0.12 &10.01&\cellcolor{lo}3.33&0.14  & 11.64 & 3.42 & 0.15\\
    {Student+Audio+Seg+Flow} &16.95&\cellcolor{lo}2.98&0.19 &8.32&3.59&\cellcolor{lm}0.12 &\cellcolor{lb}8.20&\cellcolor{lo}3.35&\cellcolor{lm}0.12  & 11.16 & \cellcolor{lo}3.31 & 0.26\\
    {Student+Audio+Flow+Depth} &18.19&3.70&0.21 &7.95&4.43&0.13 &9.80&3.71&0.14  & 11.88 & 3.95 & 0.16\\
    {Student+Audio+Seg+Depth} &17.51&3.49&0.20 &9.33&4.26&0.13 &9.07&3.61&0.13  & 11.97 & 3.79 &0.15 \\
    {Student+Seg+Flow+Depth} &18.23&3.88&0.21 &8.79&4.89&0.13 &8.93&3.44&0.13  & 11.98 & 4.07 & 0.16\\

    {Student+A+S+F+D} &16.74&4.40&0.18 &\cellcolor{lb}7.89&3.54&0.12 &9.98&4.36&0.15  & 11.53 & 4.10 & 0.15\\
    \hline
\end{tabular}
\end{table*}

\begin{table*}[!t]
  \caption{\textbf{Results for KITTI Sequences.} Model results for each of the KITTI sequences (used to compute averages in Table~\ref{tab:across}). We train various models using nuScenes, and evaluate without fine-tuning on KITTI. `w/o Filter' refers to removal of the proposed uncertainty-aware pseudo-label sample selection mechanism. Lowest three errors are highlighted within each column.  
  }
  \label{tab:leaderboard_supp}
  \centering
  \normalsize
  \resizebox{\textwidth}{!}{%
  \begin{tabular}{p{3.6cm} p{1pt} |ccc|ccc|ccc|ccc|ccc|ccc|ccc|ccc|ccc|ccc|ccc}
    \toprule
    \multirow{2}{*}{\textbf{Method}} & \multicolumn{1}{p{0.1cm}|}{}
    & \multicolumn{3}{c}{\textbf{Seq 00}}
    & \multicolumn{3}{c}{\centering \textbf{Seq 01}}
    & \multicolumn{3}{c}{\centering \textbf{Seq 02}}
    & \multicolumn{3}{c}{\centering \textbf{Seq 03}}
    & \multicolumn{3}{c}{\centering \textbf{Seq 04}}
    & \multicolumn{3}{c}{\centering \textbf{Seq 05}}
    & \multicolumn{3}{c}{\centering \textbf{Seq 06}}
    & \multicolumn{3}{c}{\centering \textbf{Seq 07}}
    & \multicolumn{3}{c}{\centering \textbf{Seq 08}}
    & \multicolumn{3}{c}{\centering \textbf{Seq 09}}
    & \multicolumn{3}{c}{\centering \textbf{Seq 10}} \\
    \cline{3-35} 
        &\multicolumn{1}{p{0.1cm}|}{}
        & $t_{rel}$ &$r_{rel}$ & $se$ 
        & $t_{rel}$ &$r_{rel}$ & $se$ 
        & $t_{rel}$ &$r_{rel}$ & $se$ 
        & $t_{rel}$ &$r_{rel}$ & $se$ 
        & $t_{rel}$ &$r_{rel}$ & $se$ 
        & $t_{rel}$ &$r_{rel}$ & $se$ 
        & $t_{rel}$ &$r_{rel}$ & $se$ 
        & $t_{rel}$ &$r_{rel}$ & $se$ 
        & $t_{rel}$ &$r_{rel}$ & $se$ 
        & $t_{rel}$ &$r_{rel}$ & $se$ 
        & $t_{rel}$ &$r_{rel}$ & $se$ \\ \hline

    Our Teacher (nuScenes)    & & 24.62 &  7.51 & 0.24 & 40.02 & 2.44 & 0.41 & 25.19 & 5.14 & 0.25 & 26.78 & 4.92 & 0.25 & 26.02 & 2.42 & 0.26 & 23.65 & 8.85 & 0.22 & 23.97 & 6.47 & 0.24 & 30.66 &20.32 & 0.26 & 23.03 & 6.69 & 0.20 & 23.23 & 4.44 & 0.22 & 20.57 & 6.01 & 0.18 \\
    \hline

    Student w/o Filter   & & 18.16 & 6.45 & 0.19 & 41.24 &\cellcolor{lo}1.84 & 0.41 & 19.05 & 4.48 & 0.22 & 26.98 & 9.68 & 0.23 & 22.56 & 2.15 & 0.22 & 14.77 & 5.83 & 0.17 & 11.38 & \cellcolor{lo}1.62 & 0.18 & 16.45 &9.35 & 0.20 & 19.36 & 7.95 & 0.18 & 16.82 & 4.14 & 0.20 & 20.23 & 8.99 & 0.15 \\

    Student   & & 14.71 & 5.42 & 0.12 & 39.79 & 2.56 & 0.38 & \cellcolor{lb}12.54 & 2.82 & 0.14 & 20.30 & 3.97 & 0.18 & 16.33 & 1.57 & 0.16 & 11.12 & 4.19 & \cellcolor{lm}0.11 & 15.60 & 5.69 & \cellcolor{lm}0.11 & \cellcolor{lb}7.77 &\cellcolor{lo}3.48 & 0.18 & 13.15 & 4.41 & 0.13 & 16.17 & 4.50 & \cellcolor{lm}0.12 & 19.91 & 5.59 & 0.14 \\

    Student+Seg w/o Filter   & &  12.41 & 3.32 & 0.17 & 41.20 & 3.28 & 0.39 & 15.70 & 2.56 & 0.19 & 22.16 & \cellcolor{lo}3.09 & 0.22 & 22.39 & 3.23 & 0.22 & 10.92 & 3.51 & 0.15 & \cellcolor{lb}10.41 &\cellcolor{lo}1.86 & 0.15 & 12.88 &6.57 & 0.19 & 10.53 & 3.14 & 0.15 & 14.14 & 2.88 & 0.16 & 10.36 & 4.21 & 0.15 \\

    Student+Seg    & &  13.88 & 5.09 & 0.12 & 40.73 & 1.94 & 0.40 & 13.22 & 3.11 & 0.14 & 17.18 & 5.14 & 0.15 & 18.40 & 1.74 & 0.18 & 11.71 & 4.27& 0.12 & 15.26 & 5.03& 0.12& \cellcolor{lb}5.97 & 3.52& 0.17& 11.79 & 3.88 & 0.12 & 15.85 & 3.77 & 0.14 & 15.40 & 4.02 & 0.15 \\

    Student+Flow w/o Filter & &17.92&6.19&0.18 &42.25&\cellcolor{lo}1.45&0.43 &20.56&5.30&0.22 &21.55&4.34&0.22 &24.03&4.52&0.23 &17.00&6.82&0.19 &13.26&3.19&0.18 &18.68&10.79&0.20 &17.07&7.33&0.17 &21.11&6.03&0.21 &11.73&6.19&0.17\\

    Student+Flow    & & \cellcolor{lb}9.52&2.68&0.15 &38.94&2.40&0.40 &13.02&2.82&0.16 &23.31&\cellcolor{lo}3.09&0.23 &19.40&1.17&0.19 &\cellcolor{lb}8.41&\cellcolor{lo}1.94&0.16 &19.60&6.61&0.13 &9.24&4.68&0.21 &8.85&2.81&0.15 &12.63&2.87&0.16 &9.72&4.00&0.15  \\

    Student+Depth w/o Filter & &20.11&7.56&0.14 &40.46&2.72&0.40 &15.38&3.32&0.17 &18.92&4.83&0.16 &21.33&2.91&0.21 &12.19&5.05&0.13 &22.01&7.49&0.14 &13.77&6.38&0.18 &13.82&4.36&0.14 &15.84&2.97&0.17 &16.86&4.26&0.16\\

    Student+Depth    & & \cellcolor{lb}7.57&\cellcolor{lo}2.13&0.13 &38.66&2.96&\cellcolor{lm}0.37 &\cellcolor{lb}11.38&2.75&\cellcolor{lm}0.13 &\cellcolor{lb}16.12&\cellcolor{lo}3.29&0.17 &\cellcolor{lb}11.71&\cellcolor{lo}0.91&\cellcolor{lm}0.12 
    &9.96&3.00&0.12 
    &13.58&5.51&0.11 
    &8.05&4.48&0.18 
    &\cellcolor{lb}8.58&2.19&0.12 
    &\cellcolor{lb}10.58&2.84&0.13 
    &\cellcolor{lb}7.84&3.27&0.15 \\

    Student+Audio w/o Filter   & &  13.01 & 4.00 & 0.17 & 39.97 & 3.58 & 0.38 & 15.71 & 3.26 & 0.18 & 21.48 & 4.80 & 0.21 & 21.15 & 2.05 & 0.21 & 10.84 & 3.83 & 0.14 & \cellcolor{lb}10.09 & \cellcolor{lo}1.50 & 0.16 & 12.69 &6.76 & \cellcolor{lm}0.17 & 12.37 & 4.39 & 0.15 & 13.76 & 3.16 & 0.16 & 9.89 & 5.37 & 0.14 \\
    
    Student+Audio    & &  \cellcolor{lb}9.09 & \cellcolor{lo}2.01 & \cellcolor{lm}0.12 & \cellcolor{lb}37.46 & 3.32 & \cellcolor{lm}0.35 & 13.87 & \cellcolor{lo}2.46 & 0.16 & \cellcolor{lb}16.06 & 3.97 & 0.15 & 19.27 & 1.78 & 0.19 & 9.77 & 2.98 & 0.13 & 10.91 & 3.16 & 0.12 & 11.26 &6.61 & \cellcolor{lm}0.17 & \cellcolor{lb}7.91 & \cellcolor{lo}1.84 & \cellcolor{lm}0.12 & 13.42 & \cellcolor{lo}2.77 & 0.14 & \cellcolor{lb}9.01 & \cellcolor{lo}2.83 & \cellcolor{lm}0.14 \\


    Student+Audio+Seg    & &  10.49 & 3.71 & \cellcolor{lm}0.12 & 42.46 & 2.73 & 0.40 & \cellcolor{lb}11.61 & \cellcolor{lo}2.38 & \cellcolor{lm}0.14 & \cellcolor{lb}14.53 & 3.93 & \cellcolor{lm}0.14 & 16.29 & \cellcolor{lo}0.96 & 0.17 & \cellcolor{lb}8.31 & \cellcolor{lo}2.76 & \cellcolor{lm}0.11 & 15.31 & 5.49 & \cellcolor{lm}0.11 & \cellcolor{lb}5.86 &\cellcolor{lo}3.00 & \cellcolor{lm}0.17 & \cellcolor{lb}7.05 & \cellcolor{lo}2.03 & \cellcolor{lm}0.12 & \cellcolor{lb}11.64 & \cellcolor{lo}2.78 & \cellcolor{lm}0.13 & 12.17 & 3.45 & \cellcolor{lm}0.14 \\



    Student+Audio+Depth    & &12.99&4.71&0.16 &37.79&\cellcolor{lo}1.63&0.39 &15.47&3.94&0.16 &17.06&3.37&0.18 &13.42&2.63&0.14 &\cellcolor{lb}9.53&2.77&0.14 &20.02&7.20&0.13 &8.95&3.70&0.19 &10.18&3.47&0.15 &15.12&4.57&0.16 &\cellcolor{lb}9.40&4.42&\cellcolor{lm}0.14\\

    Student+Seg+Depth    & &14.73&5.22&\cellcolor{lm}0.12 &39.28&2.39&\cellcolor{lm}0.37 &13.72&3.29&\cellcolor{lm}0.11 &17.98&5.28&\cellcolor{lm}0.13 &\cellcolor{lb}11.38&1.91&\cellcolor{lm}0.11 &14.84&5.76&\cellcolor{lm}0.11 &\cellcolor{lb}8.47&2.05&\cellcolor{lm}0.10 &10.68&7.20&0.19 &12.25&3.96&\cellcolor{lm}0.11 &\cellcolor{lb}10.71&3.12&\cellcolor{lm}0.10 &11.78&3.58&0.16\\

    Student+Audio+Flow & &11.55&2.23&0.18 &\cellcolor{lb}36.98&2.89&0.37 &15.22&3.09&0.18 &20.41&3.35&0.20 &19.83&2.33&0.20 &10.85&\cellcolor{lo}2.63&0.16 &12.24&2.96&0.16 &11.43&4.56&0.22 &8.83&\cellcolor{lo}2.16&0.15 &15.12&3.74&0.15 &11.78&4.08&0.15\\

    Student+Flow+Depth & &11.90&\cellcolor{lo}2.15&0.18 &41.76&2.44&0.41 &15.51&\cellcolor{lo}2.56&0.18 &22.31&3.81&0.22 &20.00&1.95&0.20 &12.58&3.16&0.16 &12.15&2.85&0.16 &14.35&6.93&0.20 &11.21&2.32&0.15 &15.95&\cellcolor{lo}2.56&0.18 &10.18&\cellcolor{lo}3.06&0.15\\

    Student+Audio+Seg+Flow & &11.51&3.19&0.15 &41.72&3.31&0.42 &15.72&2.70&0.19 &22.06&3.58&0.22 &20.36&1.76&0.20 &10.28&2.78&0.13 &13.98&3.51&0.16 &8.84&\cellcolor{lo}3.44&0.18 &10.47&2.33&0.14 &16.64&2.87&0.18 &14.85&\cellcolor{lo}3.33&0.15\\

    Student+Audio+Flow+Depth & &13.02&2.69&0.20 &42.33&2.15&0.42 &15.99&2.82&0.19 &22.36&4.39&0.23 &21.46&2.27&0.21 &12.64&3.24&0.18 &12.83&2.80&0.16 &15.04&8.07&0.21 &12.14&3.19&0.17 &18.42&3.81&0.20 &13.89&5.26&0.15\\

    Student+Audio+Seg+Depth & &13.27&2.78&0.18 &42.21&2.22&0.41 &15.79&2.81&0.19 &20.24&3.93&0.19 &24.15&3.42&0.24 &12.75&3.61&0.16 &12.86&2.78&0.15 &14.22&7.30&0.22 &10.68&2.30&0.15 &16.67&3.61&0.19 &9.78&3.61&0.16\\

    Student+Seg+Flow+Depth & &13.12&2.58&0.19 &41.47&1.87&0.41 &17.48&3.06&0.21 &20.40&3.83&0.20 &21.89&3.73&0.21 &13.61&3.87&0.18 &13.43&2.54&0.17 &15.98&8.15&0.21 &12.49&3.37&0.17 &18.39&3.90&0.21 &12.28&5.76&0.17\\

    Student+A+S+F+D & &14.98&4.59&0.16 &\cellcolor{lb}37.72&2.28&0.37 &16.23&4.57&0.14 &16.76&5.43&\cellcolor{lm}0.14  &\cellcolor{lb}12.10&\cellcolor{lo}1.17&\cellcolor{lm}0.13 &16.70&5.28&0.15 &10.80&3.59&0.17 &15.69&8.95&0.24 &13.98&4.07&0.14 &12.52&3.93&0.13 &16.66&4.55&0.17  \\
    
    \hline
    
  \end{tabular}%
}
\end{table*}

\boldparagraph{Results on All Benchmarks}  Although we do find fluctuations in the benefit of auxiliary tasks within each dataset, the audio-based model is shown to perform consistently best across all datasets (10.54 translation, 3.11 rotation, and 0.13 scale errors). While other combinations of auxiliary tasks also perform well, we do not find them to benefit generalized representation learning as much as the audio task itself. Overall, the results affirm our hypothesis that the audio task can act as a useful regularizer for the VO task. While the audio can have significant ambient noise, the pseudo-labels with the depth, flow, and segmentation tasks can also contain noise. Moreover, many approaches for such tasks are often developed and tuned for KITTI, potentially incorporating various implicit biases towards KITTI-specific settings. Nonetheless, the addition of such tasks can benefit the audio-only model at times, \eg, on KITTI with the audio+segmentation model (14.20 vs. 14.37 translation error and 3.02 vs. 3.06 rotation error). As combining multiple modalities can potentially interfere with the main VO task as well as introduce noise, effective combinations with multiple auxiliary tasks require further study in the future. 


\boldparagraph{Scale Error Discussion} Based on our analysis in Table~\ref{tab:across}, we do show the scale error to differ from the other metrics. For instance, the segmentation+depth model has the lowest scale error on KITTI but not translation nor rotation errors. Similarly, the audio+segmentation+flow model achieves low scale error on Argoverse. Moreover, while the translation error between the depth and audio-based models on KITTI varies, their scale error is similar (audio slightly lower). As the audio-based model outperforms the depth in terms of overall performance, we can see how the \textit{\textbf{scale error provides a better predictor}} in this case for overall performance. The scale error is shown to consistently select the audio-based student model on the other datasets, while translational error exhibits more variability. For instance, on Argoverse, the translation errors of audio+flow, flow+depth, and all tasks are lower than the audio model, yet the scale error selects the audio model.







\subsection{Uncertainty-Aware Pseudo-Label Filter}
\label{subsec:filter_supp}
Table~\ref{tab:across} also analyzes the role of the uncertainty-based removal of pseudo-labeled samples prior to training the student model. We find our sample removal mechanism to generally benefit the student models regardless of the task. The impact is more pronounced for the student, flow, and depth models. The segmentation model shows a slight increase in average overall translation error (from 11.60 to 12.28), but reduction in rotation and scale errors. Moreover, the scale error for the audio-based model significantly drops after filtering out potentially noisy pseudo-labels, from 0.16 to 0.13. These findings affirm the benefits of our proposed approach as well as evaluation with the scale error. 



\subsection{Additional Pseudo-Labeling Iterations}
\label{subsec:iter}
We investigated the benefit of additional iterations of pseudo-labeling, where the trained student model is used to re-label the YouTube videos with potentially higher quality pseudo-labels. However, we did not find additional pseudo-labeling iterations and re-training to benefit model performance. For instance, in the case of the strongest audio-based model, the scale error increases slightly (from 0.13 and 0.14) following another iteration. The translation and rotation errors also slightly increase by 3\% and 2\%, respectively, and plateau with subsequent iterations.

\section{Additional Qualitative Results}
\label{sec:qual}

\begin{figure*}[t]
    \centering
    \includegraphics[width=0.99\textwidth]{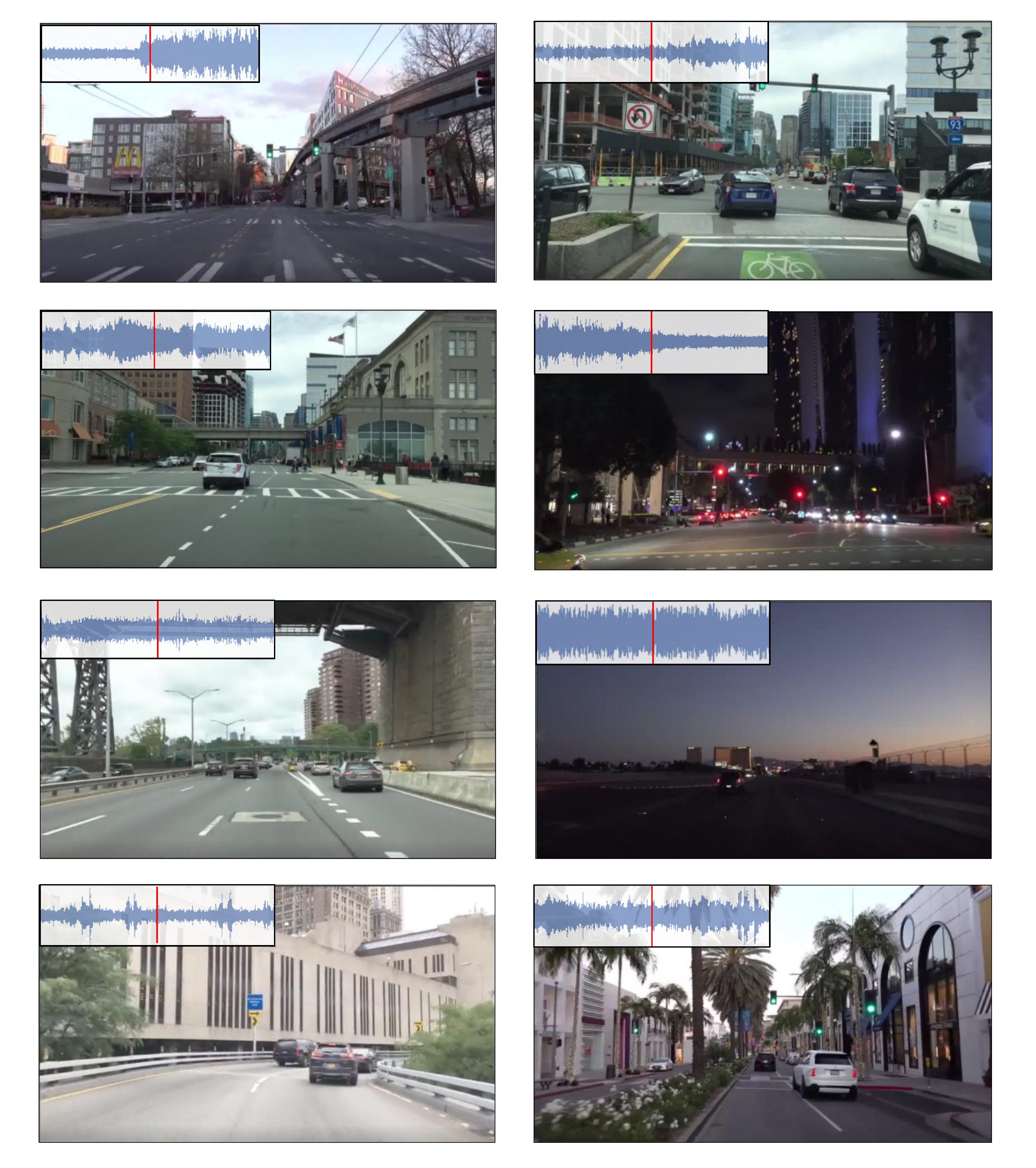}
    \caption{\textbf{Role of Audio.}
    First row: The audio amplitude is shown to increase when the vehicle speeds up from a stopped state; Second row: The audio amplitude decreases when the vehicle begins to slow down; Third row: High and stable audio amplitude on a freeway; Row 4: Various urban events, \eg, stop-and-go traffic, presence of surrounding vehicles, or slowing down to turn. }
   \label{fig:audio_supp}
\end{figure*}

\begin{figure*}[t]
    \centering
    \includegraphics[width=0.75\textwidth, trim={0cm 75cm 0cm 0cm},clip]{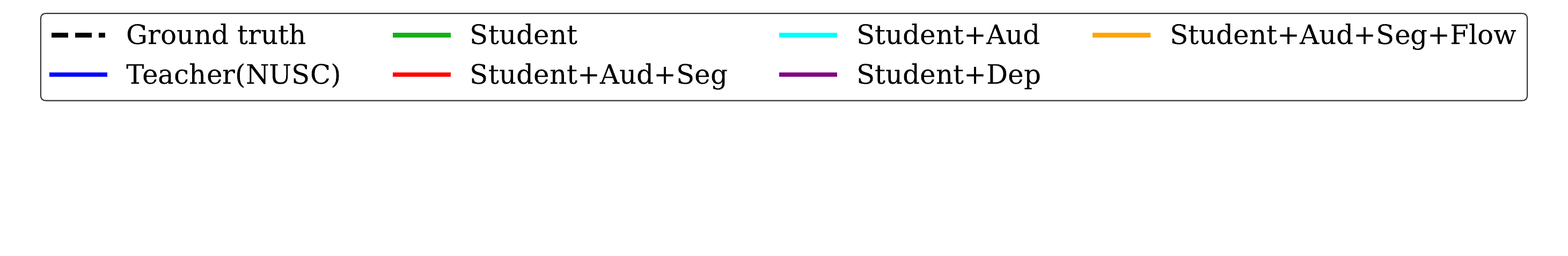} 
    \bigbreak
    \includegraphics[width=0.75\textwidth, trim={0cm 0cm 0cm 10cm},clip]{argo_v3.pdf}
    \caption{\textbf{Recovered Trajectories on Argoverse.}   First two rows depict \textbf{success cases} where the proposed approach is shown to improve the predicted trajectory. Third row (left) depicts a \textbf{failure case} due to incorrect estimation of rotation during a turn.}
    \label{fig:argo}
\end{figure*}

\begin{figure*}[t]
    \centering
    \includegraphics[width=0.75\textwidth, trim={0cm 75cm 0cm 0cm},clip]{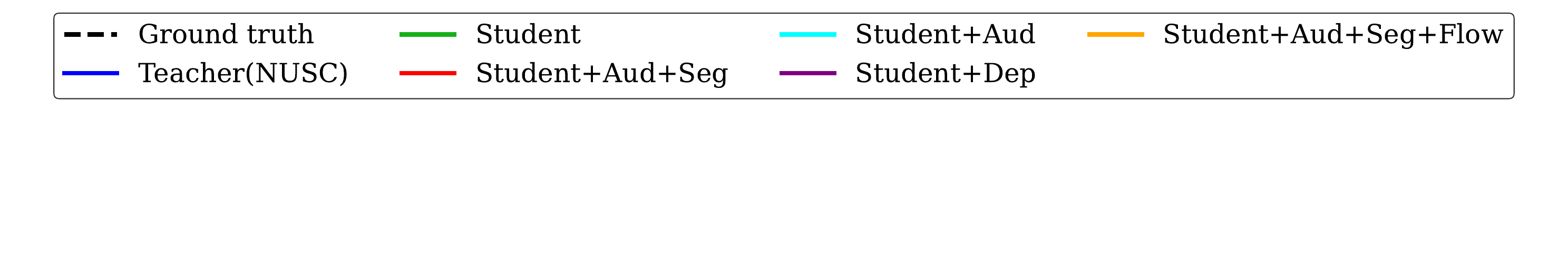} 
    \bigbreak
    \includegraphics[width=0.75\textwidth, trim={0cm 0cm 0cm 10cm},clip]{nusc_v3.pdf}
       \caption{\textbf{Recovered Trajectories on nuScenes.} First two rows depict \textbf{success cases} where the proposed approach is shown to improve the predicted trajectory. The third row (left) depicts a \textbf{failure case}, due to a challenging turn in a busy intersection where most approaches under-shoot estimated pose.  }
    \label{fig:nusc}
\end{figure*}

\begin{figure*}[t]
\centering
    \includegraphics[width=0.75\textwidth, trim={0cm 75cm 0cm 0cm},clip]{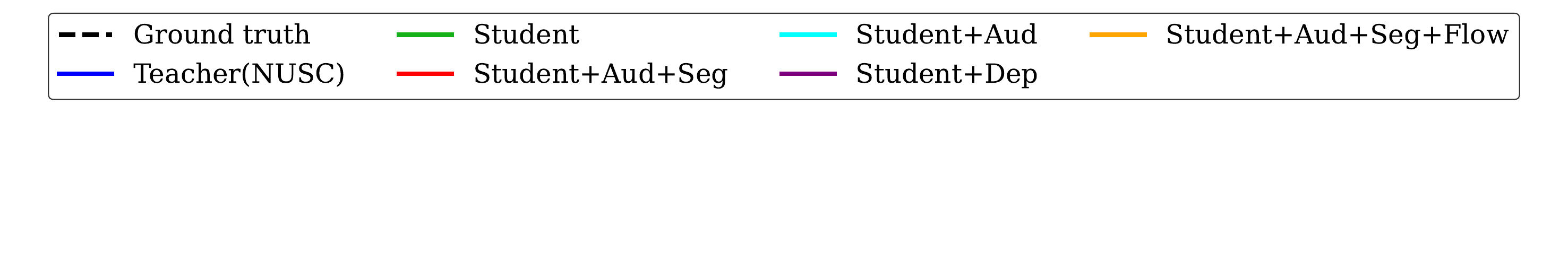} 
    \bigbreak
    \includegraphics[width=0.75\textwidth, trim={0cm 0cm 0cm 10cm},clip]{kitti_v3.pdf}
   \caption{\textbf{Recovered Trajectories on KITTI.}   First two rows depict \textbf{success cases} where the proposed approach is shown to improve the predicted trajectory. The third row depicts two \textbf{failure cases} where most of the approaches fail, either due to a lengthy forward motion on a stretch minimal clear visual landmarks or along a curved route. }
    \label{fig:kitti}
\end{figure*}

To further motivate our use of the audio modality, Fig.~\ref{fig:audio_supp} depicts additional examples from our dataset with overlaid audio. In addition to ego-motion speed (\eg, stopped or driving), the audio cues also contain general context regarding the traffic scenario (\eg, highway, urban). We also visualize additional success and failure examples for Argoverse (Fig.~\ref{fig:argo}), nuScenes (Fig.~\ref{fig:nusc}), and KITTI (Fig.~\ref{fig:kitti}). 


